\documentclass[english, 10pt, twocolumn, twoside]{IEEEtran}

\usepackage{times}
\usepackage{epsfig}
\usepackage{graphicx}
\usepackage{amsmath}
\usepackage{amssymb}
\usepackage{multirow}
\usepackage[english]{babel}
\usepackage {subcaption}
\usepackage{float}
\usepackage{cuted}
\usepackage{color}
\usepackage{hyperref}

\ifCLASSOPTIONcompsoc
\usepackage[nocompress]{cite}
\else
\usepackage{cite}
\fi

\begin{document}
	%
	\title{PoNA: Pose-guided Non-local Attention \\for Human Pose Transfer}

	\author{Kun Li$^{\dagger}$,~\IEEEmembership{Member,~IEEE,}~Jinsong Zhang$^{\dagger}$, Yebin Liu,~\IEEEmembership{Member,~IEEE,}~Yu-Kun Lai,~\IEEEmembership{Member,~IEEE,}\\ and~Qionghai Dai,~\IEEEmembership{Senior Member,~IEEE}
\thanks{This work was supported in part
by Disruptive Technology of the Ministry of Science and Technology
(2020AAA0130000), the National Key Research and Development Program of China (2018YFB2100500), and Tianjin Research Program of Application Foundation and Advanced Technology (18JCYBJC19200).}
	\thanks{$\dagger$ Equal contribution.}
	\thanks{Kun Li and Jinsong Zhang are with the College of Intelligence and Computing, Tianjin University,
Tianjin 300350, and the Key Research Center for Surface Monitoring and Analysis of Cultural Relics (SMACR), State Administration of Cultural Heritage, China.}
	 \thanks{Yebin Liu and Qionghai Dai are with the Department of Automation, Tsinghua University, Beijing 10084, China.}
	\thanks{Yu-Kun Lai is with the School of Computer Science and Informatics, Cardiff University, Cardiff CF24 3AA, United Kingdom.}
	}
	
	\markboth{}%
	{Li \MakeLowercase{\textit{et al.}}: PoNA: Pose-guided Non-local Attention for Human Pose Transfer}
	\maketitle
	

	\begin{abstract}
		Human pose transfer, which aims at transferring the appearance of a given person to a target pose, is very challenging and important in many applications. Previous work ignores the guidance of pose features or only uses local attention mechanism, leading to implausible and blurry results. We propose a new human pose transfer method using a generative adversarial network (GAN) with simplified cascaded blocks. In each block, we propose a pose-guided non-local attention (PoNA) mechanism with a long-range dependency scheme to select more important regions of image features to transfer. We also design pre-posed image-guided pose feature update and post-posed pose-guided image feature update to better utilize the pose and image features. Our network is simple, stable, and easy to train. Quantitative and qualitative results on Market-1501 and DeepFashion datasets show the efficacy and efficiency of our model. Compared with state-of-the-art methods, our model generates sharper and more realistic images with rich details, while having fewer parameters and faster speed. Furthermore, our generated images can help to alleviate data insufficiency for person re-identification.
	\end{abstract}
	



	%

	\section{Introduction}
	\label{sec:introduction}
	
\begin{figure}[!t]
		\centering
		\includegraphics[width=1.0\linewidth]{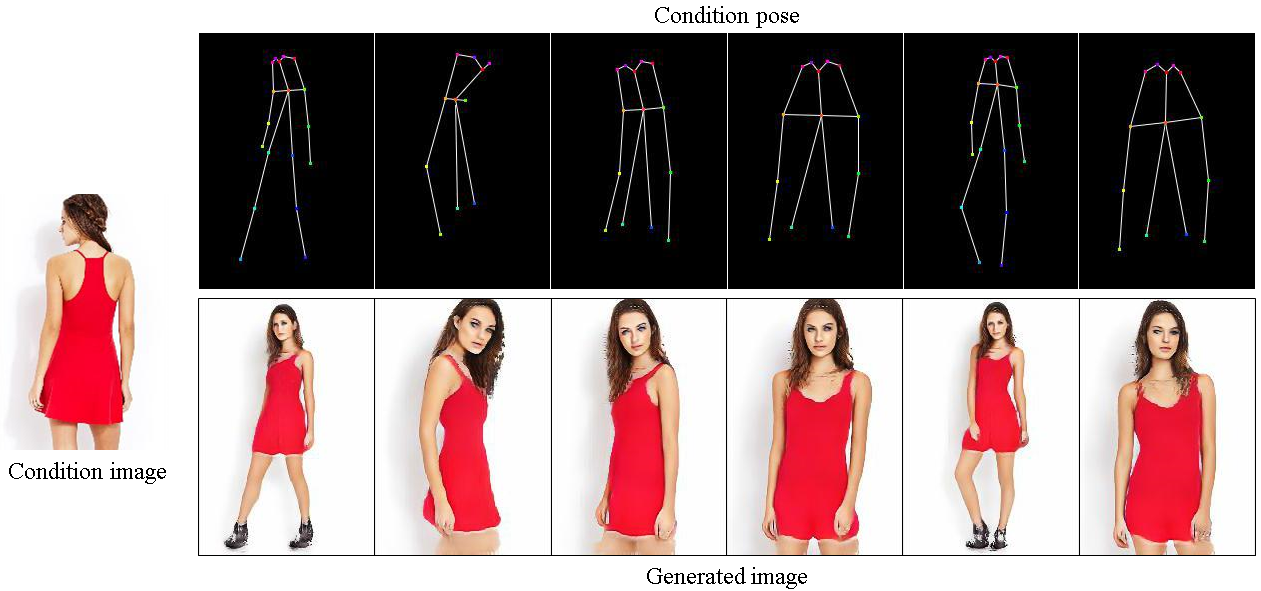}
		\caption{Generated examples by our method based on different target poses.}
\vspace{-0.6cm}
		\label{fig1}
	\end{figure}

	\IEEEPARstart{H}{uman} pose transfer, which synthesizes a new image for a target person in a new pose, is a very significant task in many applications such as data augmentation for person re-identification \cite{8709994}, image processing \cite{yue2017contrast}, and video generation \cite{Walker-2017-103533}. Given a condition image of a person and an arbitrary pose, human pose transfer system generates realistic images of the same person in the specified pose, as illustrated in Figure~\ref{fig1}.
	
	Many promising frameworks have been proposed for human pose transfer \cite{Esser_2018_CVPR,NIPS2017_6644,Ma_2018_CVPR,Siarohin_2018_CVPR,Walker-2017-103533,Li_2019_CVPR,Zhu_2019_CVPR}. In order to generate realistic images, three main ideas are used in previous work. The first kind of methods \cite{NIPS2017_6644,Esser_2018_CVPR,Ma_2018_CVPR,Zhu_2019_CVPR} apply an encoder-decoder framework to implicitly accomplish the transformation. However, they fuse the pose and image information through simple guide mechanism without utilizing the pose information for guidance, which leads to blurry and implausible results.
	The second kind of methods \cite{Siarohin_2018_CVPR} break up the whole body into parts (each body part as a rectangular region), transfer each body part respectively by computing a set of affine transformations, and finally combine the information of each body part to deliver the final results. However, the rectangular regions of image features are not precise for complex background and large pattern, which sometimes causes implausible images to be generated.
	The third kind of methods \cite{Li_2019_CVPR} propose to guide the pixel-to-pixel transfer and texture transfer by 3D prior knowledge, which can generate promising results. But their results depend on the accuracy of the appearance flow and their strategies require high computational cost and complicated training procedure.

	The aforementioned existing methods fail to synthesize photo-realistic images due to the challenges of human pose transfer.
	One challenge is how to transfer the information of the condition image, including the style of clothes and appearance of the person, from the condition pose to the target pose.
	By comparing the image information to a student and pose information to a teacher, the student should study under the guidance of the teacher, while the teacher also needs to change teaching methods according to the aptitude and feedback of the student. On the one hand, under the guidance of the teacher, the student will become stronger and stronger. On the other hand, with the feedback from the student, the teacher will adjust the way of guidance according to the state of student to better guide the progress of the student.
	In fact, this is a chicken and egg problem between pose features and image features: good pose features will help to generate good image information while good image features will contribute to extracting relevant and important pose features. In previous network architectures, they ignore the guidance function between image features and pose features or use simple attention mechanism to deal with this chicken and egg problem, which is difficult to make the utmost of image and pose features.
	Besides, human pose transfer is further compounded by self-occlusion and high variance in poses, which induce ambiguities in inferring unobserved pixels. Some methods deploy deeper networks or use 3D prior knowledge to cope with this challenge. However, their strategies require large computation budget and generate blurry images especially when there is a lot of regions to be inferred due to significant difference between poses.


    The insight that takes human pose transfer as a chicken and egg problem motivates us to design a cross-modal block, named as \textbf{Po}se-guided \textbf{N}on-local \textbf{A}ttention (PoNA) block, with pre-posed image-guided pose feature update and post-posed pose-guided image feature update to better deal with the chicken and egg problem. With simplified cascaded cross-modal blocks, the model contributes to gradually transferring image features from the condition pose to the target pose.
	In the pre-posed image-guided pose feature update, we use self-attention module to merge pose feature and image feature.
	In the post-posed pose-guided image feature update, we propose a pose-guided non-local attention mechanism to alleviate ambiguities in inferring unobserved pixels, which also helps to reduce the required number of blocks. With our non-local attention mechanism, more important regions of image features can be selected and deformed, which is useful for inferring unobserved pixels and transferring image features from the condition pose to the target pose.
	Experimental results demonstrate that our method achieves more photo-realistic human pose transfer results with fewer parameters and faster speed, compared with five state-of-the-art methods. Some examples generated by our method are shown in Fig. \ref{fig1}. The code is available for research purposes at \href{https://github.com/Zhangjinso/PoNA}{https://github.com/Zhangjinso/PoNA}.
	
	Our main contributions are summarized as follows:
	
	\begin{itemize}
		\item We propose a simple yet effective generator with simplified cascaded blocks for human pose transfer, which is easy to train with fewer parameters. We will make the code publicly available online.

		\item We propose a cross-modal block with pre-posed image-guided pose feature update and post-posed pose-guided image feature update, to better deal with the chicken and egg problem. This copes well with high variance between the source image and target image, because the pre-posed image-to-pose transfer gives a better initialization for image-based transfer, which is similar to the effect of rigging for model-based animation.
		

		\item We propose a pose-guided non-local attention mechanism in the image feature update to help select and deform important regions of image features, which deals well with the information missing and self-occlusion problems.
		
		\item We demonstrate the advantage of our method over the state-of-the-arts by quantitative and qualitative evaluation, and show the capability to alleviate data insufficiency for person re-identification.
		
	\end{itemize}
	
	\begin{figure*}[!ht]
		
		\centering
		\begin{center}
			\includegraphics[scale=0.5]{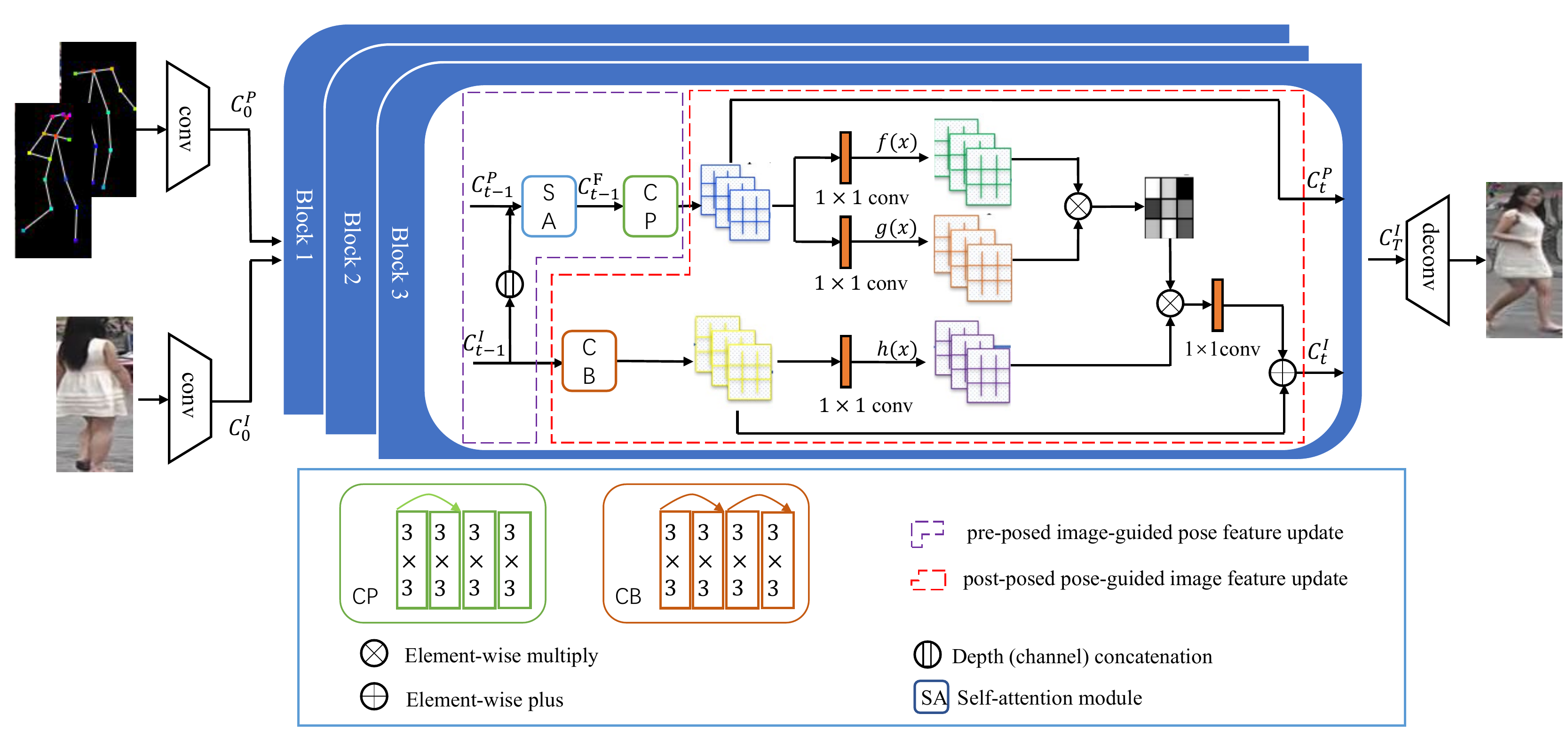}
		\end{center}
		\caption{Structure of our proposed generator.}
		\label{fig2}
	\end{figure*}
	
	The rest of this paper is organized as follows. Section  \ref{sec:related} presents a brief review of related work. Section \ref{sec:method} describes the proposed network with pose-guided non-local attention blocks. Experimental results are presented in Section \ref{sec:experiment}, and the paper is concluded in Section \ref{sec:conclude}.

	%
	%
	%
	%

	\section{Related Work}
	\label{sec:related}

	\subsection{Generative Adversarial Networks (GANs)}
	The Generative adversarial network (GAN) \cite{NIPS2014_5423} is composed of a generator and a discriminator where the discriminator tries to identify real images and synthesized images while the generator tries to fool the discriminator by generating realistic images. GANs usually generate realistic images through training in an adversarial way \cite{NIPS2014_5423, johnson2016perceptual,radford2015unsupervised}. Conditional generative adversarial networks (CGANs) \cite{Mirza2014Conditional}, which have achieved remarkable success, are proposed for the purpose of synthesizing realistic images with some condition constraints, \emph{e.g.}, generating images at new viewpoints. Besides, as a commonly used structure in generators, U-Net \cite{ronneberger2015u} captures the input information of encoder and conveys it to decoder using skip connection, which is suitable for pixel-wise aligned tasks. However, human pose transfer is an unaligned task due to the deformation between the condition pose and target pose. Self-attention GAN \cite{pmlr-v97-zhang19d} applies long-range dependency instead of local spatial dependency, solving the limitation of convolution operator. It calculates the attention map through transforming image features into two feature spaces to gain global information, and then acts on image features to get new image features with long-range dependency. However, self-attention mechanism can only gain the long-range dependency of itself but can not guide the transformation of other features, \emph{i.e.}, guiding the transformation of image features using pose features. In this paper, we propose an improved attention mechanism for human pose transfer, which helps to deal with missing information and self-occlusions.
	
	\subsection{Person Image Generation}
	Lassner \emph{et al.} \cite{Lassner:GeneratingPeople:2017} combine variational auto-encoder \cite{kingma2013auto-encoding} and GAN to generate random person images with different appearance for the full body. Xian \emph{et al.} \cite{XIAN2018CVPR} present a model for controlling the texture of synthesized image with input sketch and texture patches. Zhu \emph{et al.} \cite{Zhu2018CVPR} propose a novel pipeline to synthesize novel views of human body from a monocular image. Balakrishnan \emph{et al.} \cite{Balakrishnan2018CVPR} decompose the person image generation task into multiple foregrounds with different body parts and background generation. Si \emph{et al.} \cite{Si_2018_CVPR} adopt multi-stage adversarial losses for pose transformer network, foreground transformer network and background transformer network to generate more realistic images. Several methods~\cite{Han_2018_CVPR,lahner2018deepwrinkles,wang2018toward} focus on the virtual try-on application and make remarkable progress in transferring clothes of a given person image while containing the condition pose and shape of that person.
	Previous existing researches ignore the chicken and egg problem in conditional generation. In this paper, we propose a cross-modal block with pre-posed image-guided pose feature update and post-posed pose-guided image feature update to cope with the chicken and egg problem in human pose transfer.
	
	\subsection{Human Pose Transfer}
	Human pose transfer is an important part of person image generation. Ma \emph{et al.} \cite{NIPS2017_6644} propose a coarse-to-fine framework, which first generates a coarse image and then refines it. However, this two-stage model is inefficient in computation and complex to train. Ma \emph{et al.} \cite{Ma_2018_CVPR} improve their previous work using a decomposition strategy. Esser \emph{et al.} \cite{Esser_2018_CVPR} combine a variational auto-encoder to sample appearance and a U-Net \cite{ronneberger2015u} to preserve shape information, modeling the interplay of shape and appearance. Neverova \emph{et al.} \cite{neverova2018dense} propose to form a warping module to preserve texture and a prediction module to generate plausible images and then use a blending module to deliver final results. Their work exploits a dense pose estimation system, which maps body pixels to UV surface coordinates, to generate pose presentation. Li \emph{et al.} \cite{Li_2019_CVPR} use a 3D flow map and a visibility map from the condition pose and target pose to guide the transformation of image features and pixels. However, they need a flow regression module pre-trained by the dataset obtained through fitting a SMPL \cite{SMPL:2015}, a 3D human model with 6890 vertices and 13766 faces.
	Liu \emph{et al.} \cite{lwb2019} use a 3D body mesh recovery module \cite{kanazawaHMR18} to disentangle the pose and shape, and propose a novel block to propagate the source information. However, the 3D body mesh recovery method cannot estimate pose and shape precisely, which leads to some blurry results.
	Zhu \emph{et al.} \cite{Zhu_2019_CVPR} propose a progressive pose attention transfer network to utilize pose features to guide the image features transfer. They consider the guidance role of pose features, but local attention mechanism cannot capture long-range dependency to transfer the precise regions of the image features. In this paper, we cascade several pose-guided non-local attention blocks for better pose transfer using fewer parameters.

	\section{Method}
	\label{sec:method}
	
	\subsection{Notations}
	Given a person image, we aim at generating an image for the person in another pose. To transfer pose arbitrarily, we adopt the commonly used pose representations to guide the transfer. Specifically, we use 18 human keypoints extracted by Human Pose Estimator (HPE) \cite{Cao_2017_CVPR} and represented by heatmaps. The heatmap includes 18 channels, and each channel encodes the location of each joint of human body. Denote $P_c$ and $P_t$ as the heatmaps of the conditional pose and the target pose, $I_c$ and $I_t$ as the condition image and the target image.

	Instead of directly using the keypoints
	$K_c$ and $K_t$ that are extracted from the condition image $I_c$ and the target image $I_t$ using HPE, we encode the pose as 18 heatmaps to provide widespread information about each joint location. Specifically, the condition pose $P_c$ and the target pose $P_t$ are represented as two tensors $P_c$ = $\mathcal{P}(K_c)$ and $P_t$ = $\mathcal{P}(K_t)$, where the slice $\mathcal{P}_i$ (1 $\leq{i}\leq$ 18) is a 2D matrix with the same dimension as $I_c$ and $I_t$. Mathematically,
	\begin{equation}
	\mathcal{P}_{i} = \exp{-\frac{(k-k_i)^2}{\sigma^2}},
	\end{equation}
	where $k_i$ is the $i$-th joint location and $\sigma$ = 6 pixels, which is the same as that in Deform\cite{Siarohin_2018_CVPR} and PATN\cite{Zhu_2019_CVPR}.
	During training, the model takes a pair of the condition and the target images ($I_c, I_p$) and a pair of the condition and the target pose ($P_c,P_t$) as inputs. The generator is fed with a triplet ($I_c, P_c, P_t$) and outputs $I_g$ as close as possible to $I_t$.

	\subsection{Encoders}
	Figure~\ref{fig2} shows the architecture of our generator. The generator has two encoders, several pose-guided non-local attention blocks and one decoder. The input of the generator is ($I_c$, $P_c$, $P_t$), a triplet of the condition image, condition pose and target pose. The output is our generated image $I_g$, which is expected to be similar to the target image $I_t$. The task is transferring the information in condition image $I_c$, including texture, body shape, clothes style, \emph{etc}., from condition pose $P_c$ to target pose $P_t$.
	The encoders, pose encoder and appearance encoder, have the same structure with two down-sampling convolution layers. We concatenate condition pose and target pose as the input of the pose encoder. The pose encoder encodes the transformation between condition pose and target pose and preserves the information of both. The appearance encoder takes a condition image as input and encodes the information of the condition image.  After going through two encoders, pose code $C_0^P$ and image code $C_0^I$ are obtained.
	
	\subsection{Pose-guided Non-local Attention Block}
	We propose pose-guided non-local attention (PoNA) blocks, which are cross-modal blocks, to make pose features truly guide the transformation of image features. With several PoNA blocks, image features can be transferred by pose features from the condition pose to the target pose gradually. Each PoNA block is separate and has the same structure. The inputs of the ${t}^{st}$ PoNA block are image code $C_{t-1}^I$ and pose code $C_{t-1}^P$ from the $(t-1)^{st}$ PoNA block, and the outputs are image code $C_t^I$ and pose code $C_t^P$. Several PoNA blocks finally output the final image code $C_T^I$ which is put into the decoder to generate the final image, while the final pose code $C_T^P$ is discarded.
	
	As explained before, PoNA block is used to make pose features guide the transformation of image features. In each PoNA block, two pathways, image pathway and pose pathway, are designed for pose code and image code, respectively. Image pathway and pose pathway are connected with our improved non-local attention mechanism, which can accomplish the task of pose-guided transformation. As shown in Figure \ref{fig2}, image code $C_{t-1}^I$ and pose code $C_{t-1}^P$ from the ${t-1}^{st}$ PoNA block are first concatenated along the depth (channel) axis to get the fusion code. Pose code is updated with the fusion code as $C_{t}^P$, which contains both image features and pose features, and then attention map is computed using the updated pose code. Besides, image code is updated by going through four convolution layers. To get the final image code $C_{t}^I$, the attention map is modulated on the updated image features, which accomplishes the transformation of image features guided by pose features. Details about PoNA block is described in the following.
	
	\noindent
	\textbf{Image-guided Pose Code Update. }
	Before going through image pathway and pose pathway, we concatenate the image code and the pose code along the depth (channel) dimension as fusion features to make pose features know the information about transformation of image features. With fusion code, we deploy a self-attention module \cite{pmlr-v97-zhang19d} to better integrate fusion features and select more important regions for the guidance. Mathematically:
	\begin{equation}
		C_{t-1}^F = self\_attention(C_{t-1}^P||C_{t-1}^I).
	\end{equation}
	where $||$ is cited as concatenation along the depth (channel) dimension.
	
	For pose pathway, a block with four convolution layers is used to encode the information of fusion features and prepare for the following guidance. The four convolution layers (each layer with a normalization layer \cite{batchnorm2015, ulyanov2016instance} and a ReLU \cite{maas2013rectifier}) help the pose code know about the transformation of image features, which benefits for the following guidance. These four convolution layers are capable of extracting useful features from the fusion code. One of these layers also reduces the number of channels to half, making the size of the output equal to the input. The pose code is updated by:
	\begin{equation}
		C_{t}^P = conv_P(C_{t-1}^F).
	\end{equation}
	
	\noindent
	\textbf{Pose-guided Non-local Attention. }The pose transfer is to move patches from the condition pose to the target pose and to deal with the relationship between different patches. From this point of view, the pose guides the transfer by finding where to extract condition patches and where to put target patches and meanwhile maintaining the relationship between patches. In our PoNA block, such transformation is realized by the attention map denoted as {$\alpha_t$}, which are values between 0 and 1 calculated by softmax, indicating the importance of every element in the updated pose code and the extent to which the model attends to one location when synthesizing other locations.
	

	Traditional non-local attention mechanism embeds query, key and value from the same feature, calculates the attention map by computing the similarity between key and query, and updates the value with this attention map to select more important regions. For the guided attention mechanism, key and value are embedded from the feature to be guided, and query is embedded from the feature as guidance. However, in our task, the pose feature and the image feature are in the different latent spaces, and it is hard to compute similarity between image feature and pose feature. To get a reliable attention map, we embed key and query from the updated pose feature.
	The map {$\alpha_t$} is calculated from the pose code $C_{t-1}^P$. The pose code $C_{t-1}^P$ is transformed into two feature spaces, key and query space, by two different $1\times 1$ convolutional layers, denoted as mappings $f$ and $g$, where $f(x)=W_f x$ and $g(x)=W_g x$. Mathematically:
	\begin{equation}
		\alpha_{t,j,i}=softmax(m_{t,i,j}),
	\end{equation}
	where $\alpha_{t,j,i}$ indicates the relationship between the $i^{th}$ location and the $j^{th}$ location in the $t^{th}$ PoNA block, and $m_{t,i,j}$ is calculated as
	\begin{equation}
		m_{t,i,j}={f({C_{t}^P}_i)}^T\cdot{g({C_{t}^P}_j)}.
	\end{equation}
	
	\noindent
	\textbf{Pose-guided Image Code Update. }Image code is updated by going through four convolution layers and embedded into value space by a  $1\times 1$ convolutional layer. The attention map $\alpha_{t}$ from the pose-guided non-local attention mechanism can select important regions from the image code and  deform the original image code by rearranging the image feature.
	The output of the attention layer is $o_t=(o_{t,1},o_{t,2},\ldots,o_{t,j},\ldots,o_{t,N})$, where
	\begin{equation}
		\begin{split}
			o_{t} \quad &=\sum_{i=1}^N\alpha_{t,j,i}W_t(conv_I({C_{t-1}^I})_i),\\
			h(x_i )&=W_h x_i.
		\end{split}
	\end{equation}
	In the above formulation,
	$N$ is the number of locations of the features from the previous hidden layer, and
	$W_h$ and $W_t$ are the learned weight matrices, which are implemented as  1$\times$1 convolution layers.
	
	In addition, we combine the output of the attention layer with the input image code by a learnable parameter $\gamma$ which is initialized as 0. Therefore, the final output is given by
	\begin{equation}
		C_t^I=\gamma o_t+conv_I(C_{t-1}^I).
	\end{equation}
	The learnable coefficient $\gamma$ enables the block first to rely on local features and then gradually to learn to combine the non-local evidences.

	\subsection{Discriminator}
	We adopt two discriminators, appearance discriminator $D_A$ and pose discriminator $D_P$, to judge how likely $I_g$ contains the same person as $I_c$ (appearance consistency) and how well $I_g$ aligns with the target pose $P_t$ (pose consistency). The two discriminators are similar in structures, where the inputs of them are $I_g$ concatenated with either condition image $I_c$ or target pose $P_t$ along the depth axis. These inputs go through a convolutional layer (with normalization and ReLU after it) and several residual blocks (with self-attention module) after down-sampling. The outputs of the discriminators are the appearance consistency score $S^A$ and the pose consistency score $S^P$, calculated by softmax layers.
	
	\subsection{Loss Function}
	Previous methods on human pose transfer\cite{NIPS2017_6644,Ma_2018_CVPR,Siarohin_2018_CVPR,Esser_2018_CVPR,Zhu_2019_CVPR,Li_2019_CVPR} utilize multiple loss functions to supervise the training process. In this work, we adopt a combination of three loss functions as the full loss function $\mathcal{L}_{full}$, including a conditional adversarial loss $\mathcal{L}_{CGAN}$, an L1 loss $\mathcal{L}_{L1}$ and a perceptual loss $\mathcal{L}_{percep}$, which are described in details as follows.
	\par
	The full loss function is defined as
	\begin{equation}
		\mathcal{L}_{full}=\lambda_1\mathcal{L}_{CGAN}+\lambda_2\mathcal{L}_{L1}+\lambda_3\mathcal{L}_{percep},
	\end{equation}
	where $\lambda_1$, $\lambda_2$, $\lambda_3$ represent the weights of $\mathcal{L}_{CGAN}$, $\mathcal{L}_{L1}$, $\mathcal{L}_{percep}$ that contribute to $\mathcal{L}_{full}$, respectively.
	\vspace{0.2cm}
	
	\noindent
	\textbf{Conditional adversarial loss.}~The conditional adversarial loss is defined as
	\begin{small}
		\begin{equation}
			\begin{split}
				&\mathcal{L}_{CGAN}=E_{P_t\in \mathcal{P},(I_c,I_t)\in \mathcal{I}}\{\log[D_A(I_c,I_t)\cdot D_P(I_t,P_t)]\}+  \\
				&E_{P_t\in \mathcal{P},I_c\in \mathcal{I},I_g\in\hat{\mathcal{I}}}\{\log[(1-D_A(I_c,I_g))\cdot (1-D_P(I_g,P_t))]\},
			\end{split}	
		\end{equation}
	\end{small}where $\mathcal{P}$, $\mathcal{I}$ and $\hat{\mathcal{I}}$ denote the distributions of person poses, real person images and fake person images, respectively.
	
	\noindent
	\textbf{L1 loss.}~$\mathcal{L}_{L_1}$ loss represents the pixel-wise differences between the generated image $I_g$  and the ground truth $I_t$, which is defined as
	\begin{equation}
		\mathcal{L}_{\mathcal{L}1}=||I_g-I_t||_1.
	\end{equation}

	\noindent
	\textbf{Perceptual loss.}~Previous work \cite{NIPS2017_6644,Ma_2018_CVPR,Esser_2018_CVPR,Siarohin_2018_CVPR,Zhu_2019_CVPR,Li_2019_CVPR} shows that L1-distance between feature maps extracted from two images by a pre-trained CNN could make the generated images look more natural and reduce the structural differences, which performs well in style transfer\cite{johnson2016perceptual} and pose transfer\cite{NIPS2017_6644,Ma_2018_CVPR,Siarohin_2018_CVPR,Li_2019_CVPR}. The perceptual loss is defined as
	\begin{equation}
		\mathcal{L}_{percep}=\sum_{i=1}^{C_\rho}||\phi_\rho(I_g)_i-\phi_\rho(I_t)_i||_1,
	\end{equation}
	where $\phi_\rho$ denotes the outputs of the $\rho^{th}$ layer from the VGG-19 model\cite{simonyan2014very} pre-trained on ImageNet\cite{russakovsky2015imagenet}, and $\phi_\rho(\cdot)_i$ denotes the $i^{th}$ feature map of the outputs of $\phi_\rho$.

	In practice, we find that it is good enough to use the features from $Conv1\_2$ of VGG-19.

	\begin{table*}[htbp]
	\renewcommand{\arraystretch}{1.3}
\small%
	\setlength{\tabcolsep}{2.5mm}
	\begin{center}
		\caption{Quantitative comparison with five state-of-the-art methods on Market-1501 and DeepFashion.}\label{tab:tab_com}
		\begin{tabular}{|c|c|c|c|c|c|c|c|c|c|c|}
			\hline
			\multirow{2}*{Model} &\multicolumn{6}{|c|}{Market-1501}&\multicolumn{4}{|c|}{DeepFashion}\\ \cline{2-11}
			{}&SSIM & IS &mask-SSIM & mask-IS  & PSNR & PCKh & SSIM & IS &  PSNR & PCKh \\
			
			\hline
			PG$^2$ \cite{NIPS2017_6644} & 0.261 & \textbf{3.495} & 0.782 & 3.367  & 28.135 & 0.73 & 0.773 & 3.163  & 30.785 &0.89\\
			VUnet \cite{Esser_2018_CVPR}& 0.266 & 2.965 & 0.793 & 3.549  & 28.071 &0.92 & 0.763 & 3.440 &  30.856 & 0.93\\
			Deform \cite{Siarohin_2018_CVPR}& 0.291 & 3.230 & 0.807 & 3.502 & 28.227 &0.94 & 0.760 & 3.362 &  31.022& 0.94\\
			PATN \cite{Zhu_2019_CVPR} & 0.311 & 3.323 & 0.811 & 3.773  & 28.228 &0.94 & 0.773 & 3.209 &  31.160 & \textbf{0.96}\\
			LWG \cite{lwb2019} & - & - & -  & - & - & - & 0.696 & \textbf{3.478} &  28.396& 0.80\\
			\hline
			Ours (3 PoNA blocks) & \textbf{0.315} & 3.487 & \textbf{0.814} & \textbf{3.867} & \textbf{28.257}&\textbf{0.94} &\textbf{0.775} & 3.338 & \textbf{31.382}& 0.95  \\
			
			\hline
			Real Data & 1.000 & 3.890 & 1.000 & 3.706  & $\infty$ & 1.00 & 1.000 & 4.053 & $\infty$ &1.00\\
			\hline
		\end{tabular}
	\end{center}
\end{table*}

	\subsection{Training Procedures}
	We train the generator and the discriminators alternately. When training, the input of the generator is a triplet ($I_c$, $P_c$, $P_t$), and the output is a generated image $I_g$ which has the same pose as the target image $I_t$. Specifically, $I_c$ is fed to the image stream and ($P_c$, $P_t$) are fed to the pose stream. To train the discriminators, the appearance discriminator $D_A$ takes ($I_c$, $I_t$) and ($I_c$, $I_g$) as inputs to calculate the appearance consistency score $S^A$, and the pose discriminator $D_p$ takes ($P_t$, $I_t$) and ($P_t$, $I_g$) as inputs to calculate the pose consistency score $S^P$.
	
	\subsection{Implementation Details}
	We use 3 PoNA blocks in the generator. For images from Market-1501 dataset that are low resolution, we apply batch normalization\cite{batchnorm2015} in the generator for all the normalization layers. Instance normalization\cite{ulyanov2016instance} which is a better choice for the transfer task, is applied for DeepFashion dataset. The coefficients in the loss function ($\lambda_1,\lambda_2,\lambda_3$) are set to be (5,10,10) for Market-1501 dataset and (5,1,1) for DeepFashion dataset. Leaky-ReLU\cite{maas2013rectifier} is adopted after each convolution layer or normalization layer in the discriminators, and its negative slope coefficient is set to be 0.2. For all models, we adopt the Adam optimizer \cite{kingma2014adam} with $\beta_1=0.5$ and $\beta_2=0.999$ to train for 90k iterations.

	\section{Experimental Results}
	\label{sec:experiment}
	

	In this section, we first introduce the datasets and the metrics in Section \ref{sec:dataset}, and compare our method with the state-of-the-art methods in Section \ref{sec:compare}, and then we perform an ablation evaluation to study the importance of the different components of our approach in Section  \ref{sec:ablation}.
	We show the application on data augmentation for person re-identification which helps to improve the performance in Section \ref{sec:aug} and finally analyze the limitations of our model in Section \ref{sec:limitation}.
	
	\subsection{Datasets and Metrics}
	\label{sec:dataset}
	\subsubsection{Datasets} Person re-identification dataset Market-1501 \cite{zheng2015scalable} and  the In-shop Clothes Retrieval Benchmark DeepFashion \cite{liu2016deepfashion} are commonly used datasets for the evaluation of human pose transfer. Images in Market-1501 are low resolution (128$\times$64) with large variation in pose and background, which makes human pose transfer more challenging. Compared with Market-1501, the images in DeepFashion are higher resolution (256$\times$176) with clean background. Zhu \emph{et al.} \cite{Zhu_2019_CVPR} use HPE \cite{cao2017realtime} as pose joints detector and collect 263632 training pairs and 12000 testing pairs for Market-1501 and 101966 training pairs and 8570 testing pairs for DeepFashion, when proposing their PATN method. In both of these datasets, the person identities in the test set are different from those in the training set. In order to ensure the fairness of the comparison results, we adopt the training pairs and the testing pairs used in PATN \cite{Zhu_2019_CVPR} for both datasets.
	
	\begin{figure*}[!ht]
		\begin{center}
			\includegraphics[width=0.7\linewidth]{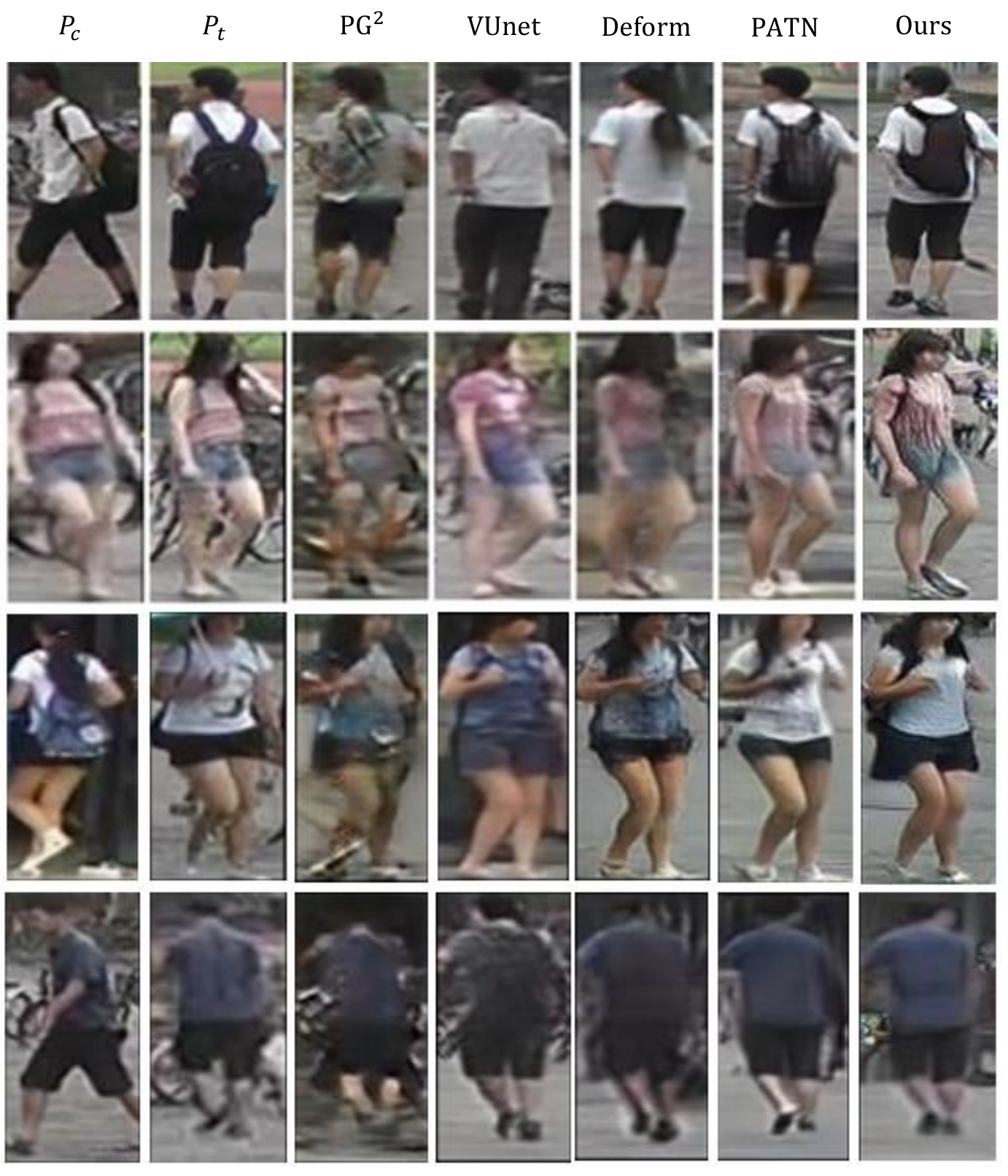}
		\end{center}
		\caption{Qualitative comparisons on Market-1501 dataset. From left to right are the results of PG$^2$~\cite{NIPS2017_6644}, VUnet~\cite{Esser_2018_CVPR}, Deform~\cite{Siarohin_2018_CVPR}, PATN~\cite{Zhu_2019_CVPR} and ours, respectively.}
		\label{fig:m_com}
	\end{figure*}
	
	\begin{figure*} [!t]
		\begin{center}
			\begin{tabular}{c} 
				\includegraphics[width=0.8\linewidth]{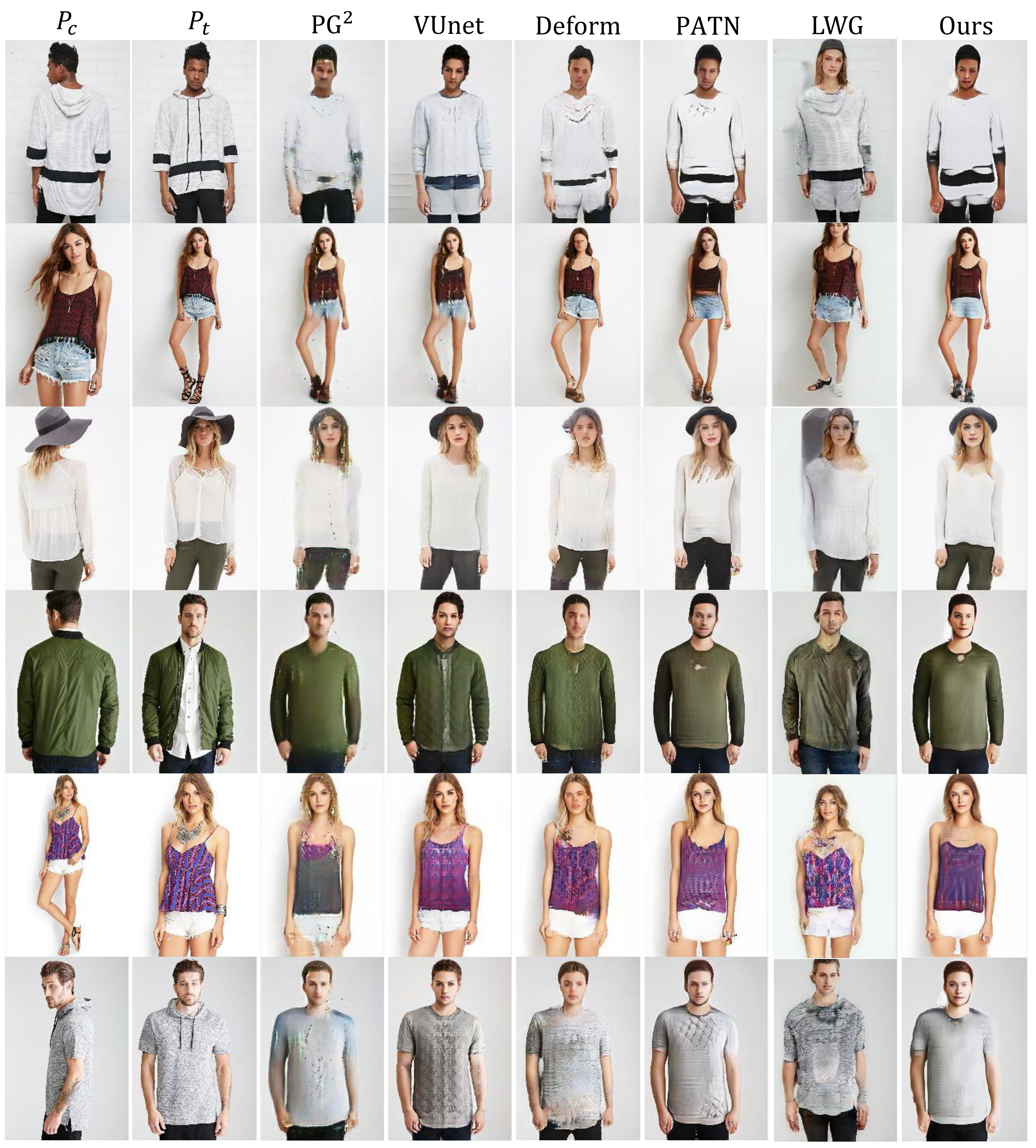}
			\end{tabular}
		\end{center}
		\caption{Qualitative comparisons on DeepFashion dataset. From left to right are the results of PG$^2$\cite{NIPS2017_6644}, VUnet\cite{Esser_2018_CVPR}, Deform\cite{Siarohin_2018_CVPR}, PATN\cite{Zhu_2019_CVPR}, LWG\cite{lwb2019} and ours, respectively.}
		\label{fig:f_com}
	\end{figure*}
	
	\subsubsection{Metrics} Ma \emph{et al.} \cite{NIPS2017_6644} use Structural SIMilarity (SSIM) and Inception Score (IS) as their metrics. SSIM can measure the similarity between synthesized images and ground-truth images, and IS is a common method used to measure the quality of image generation. For Market-1501 dataset, Ma \emph{et al.} \cite{Ma_2018_CVPR} further introduce the mask version of SSIM (mask-SSIM) and the mask version of IS (mask-IS) by masking the background out, which reduces the influence of blurry background of images. Besides, Zhu \emph{et al.} \cite{Zhu_2019_CVPR} introduce the slightly modified version of Percentage of Correct Keypoints (PCKh) \cite{Andriluka_2014_CVPR}, which measures shape consistency by computing pose joints alignment using pre-trained Human Pose Estimator (HPE) \cite{Cao_2017_CVPR}. Moreover, we use Peak Signal to Noise Ratio (PSNR) to measure the difference between the synthesized image and the ground-truth image in pixel level.
	
		\begin{figure*} [!t]
		\begin{center}
			\begin{tabular}{c} 
				\includegraphics[width=0.9\linewidth]{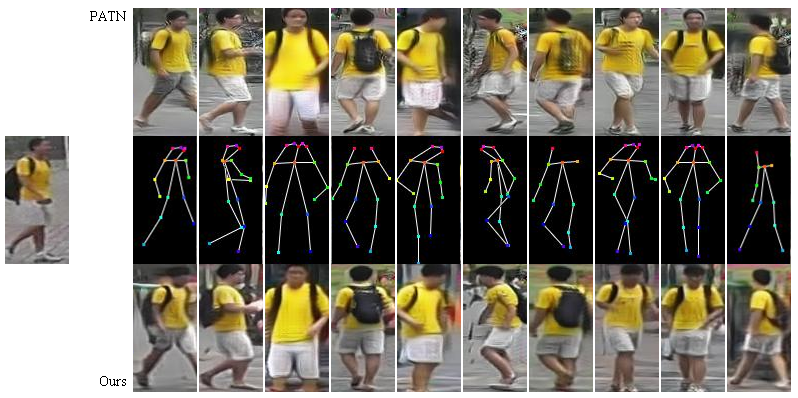}
			\end{tabular}
		\end{center}
		\caption{Image generation results conditioned by different poses on the Market-1501 dataset. For each condition image, the first row shows the images generated by PATN\cite{Zhu_2019_CVPR}, and the second row shows our results.}
		\label{fig:m_arb}
	\end{figure*}

	\subsection{Comparison with Previous Work}
	\label{sec:compare}
	\subsubsection{Qualitative Evaluation} We evaluate the visual results of our method on Market-1501 and DeepFashion datasets, compared with five state-of-the-art methods: PG$^2$ \cite{NIPS2017_6644}, VUnet \cite{Esser_2018_CVPR}, Deform \cite{Siarohin_2018_CVPR}, PATN \cite{Zhu_2019_CVPR}, and LWG \cite{lwb2019}. Because LWG \cite{lwb2019} does not provide pre-trained model on the Market-1501 dataset, we compare with LWG \cite{lwb2019} only on the DeepFashion dataset.

	For poor quality images in Market-1501, as shown in Figure \ref{fig:m_com}, our method generates clearer images than the other methods. It is worth noting that our method gives the correct leg layouts even when the legs are crossed in the target pose (in the first and third rows). Besides, even if the condition image is blurred (in the second row) or has complex clothing patterns (in the third row), our method can learn the style of the garments and maintain these features in the generated images. Moreover, our method also keeps appearance consistency, \emph{e.g.}, the bag is preserved in our results (in the first and third rows). For high quality images in DeepFashion, as shown in Figure \ref{fig:f_com}, our method yields the sharpest person images with better facial details while the generated images of the other methods have some blur contents (in the third and fifth rows). Besides, the texture (in the first and third rows) and clothing styles (in the first, second and fifth rows) in condition images are preserved in our generated images, which indicates that our model has the power of capturing global styles and local details for generation. Although LWG preserves fine details in the condition image (in the fifth row), it fails to generate sharp and plausible images with precise pose. Moreover, our method keeps appearance consistency, \emph{e.g.}, the hat in the third row.


		\begin{figure*} [!t]
\centering
			\begin{subfigure}[t]{0.9\textwidth}
				\centering
				\includegraphics[width=0.9\linewidth]{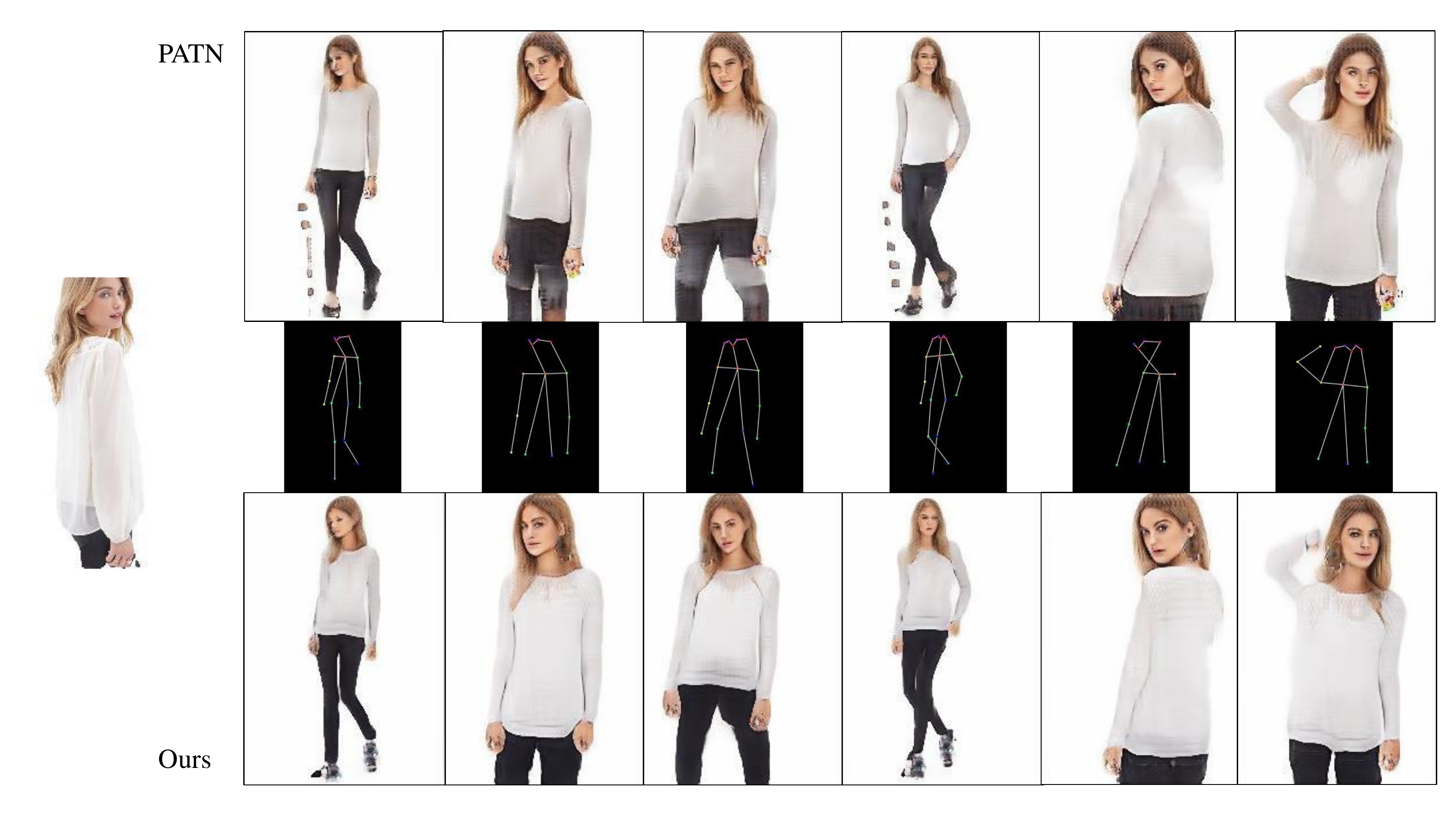}
			\end{subfigure}
		\quad
			\begin{subfigure}[t]{0.9\textwidth}
				\centering
				\includegraphics[width=0.9\linewidth]{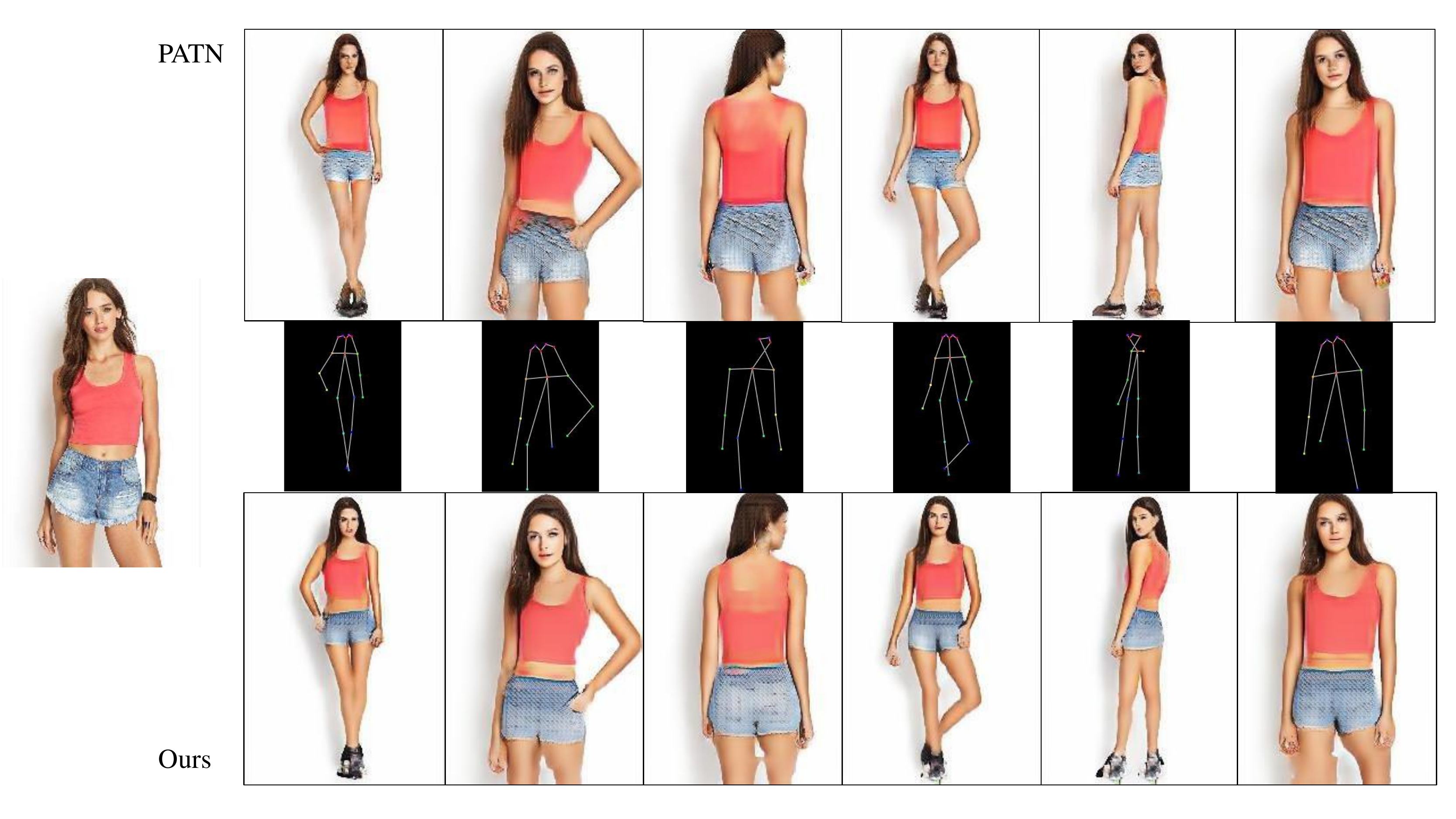}
			\end{subfigure}
			\caption{Image generation results conditioned by different poses on the DeepFashion dataset. For each condition image, the first row shows the images generated by PATN\cite{Zhu_2019_CVPR}, and the second row shows our results.}
			\label{fig:vspatn}
		\end{figure*}


		\begin{figure*} [!t]
			\centering
			\begin{subfigure}[t]{0.9\textwidth}
				\centering
				\includegraphics[width=0.9\linewidth]{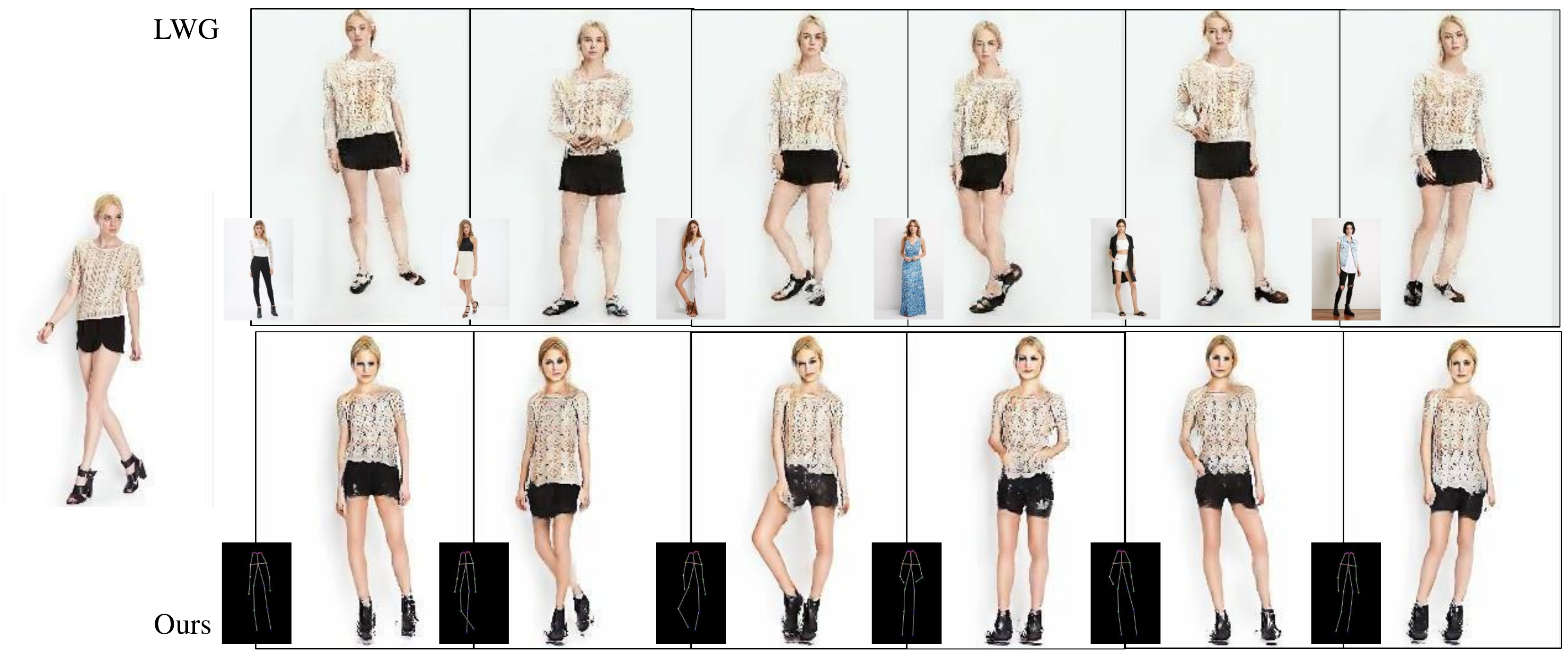}
			\end{subfigure}
			\quad
			\begin{subfigure}[t]{0.9\textwidth}
				\centering
				\includegraphics[width=0.9\linewidth]{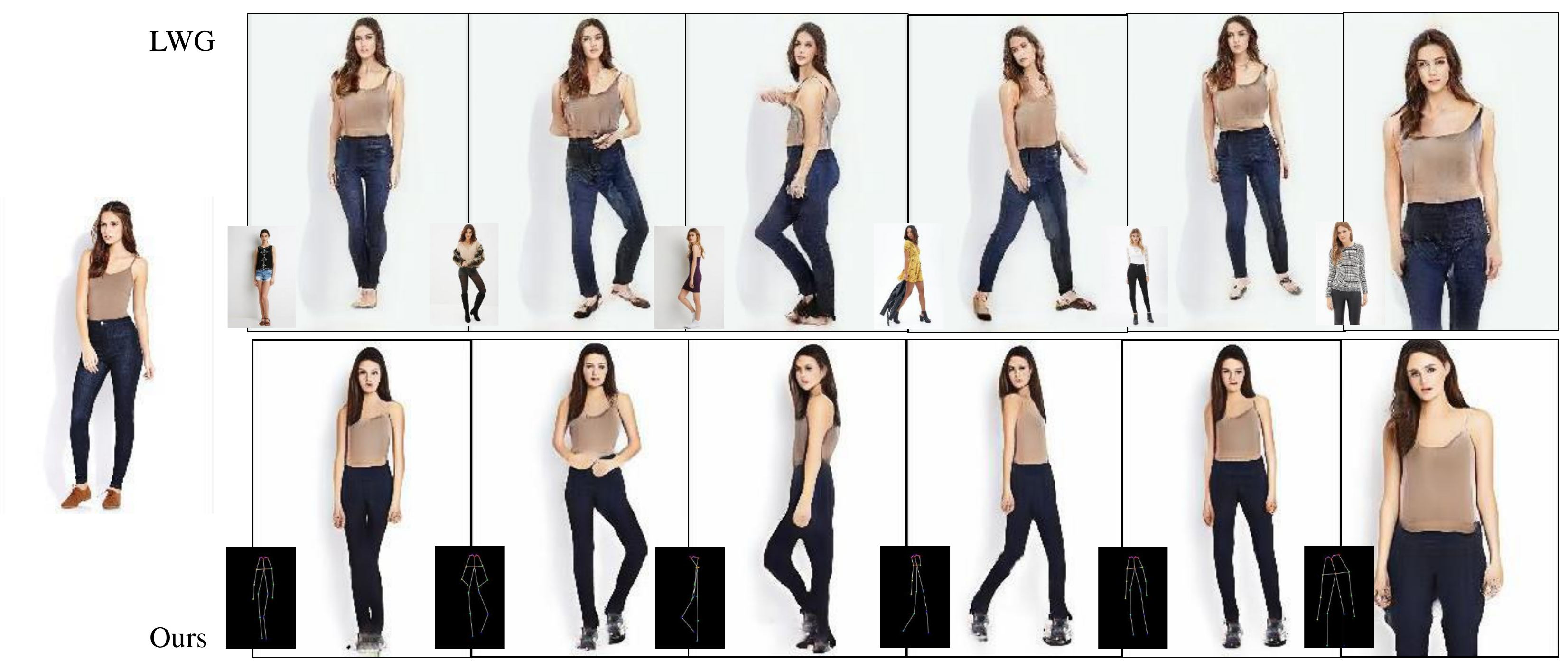}
			\end{subfigure}
			\caption{Image generation results conditioned by different poses on the DeepFashion dataset. For each condition image, the first row shows the images generated by LWG \cite{lwb2019}, and the second row shows our results.}
			\label{fig:vslwg1}
		\end{figure*}
	
	\begin{figure*} [!t]
\centering
		\includegraphics[width=0.8\linewidth]{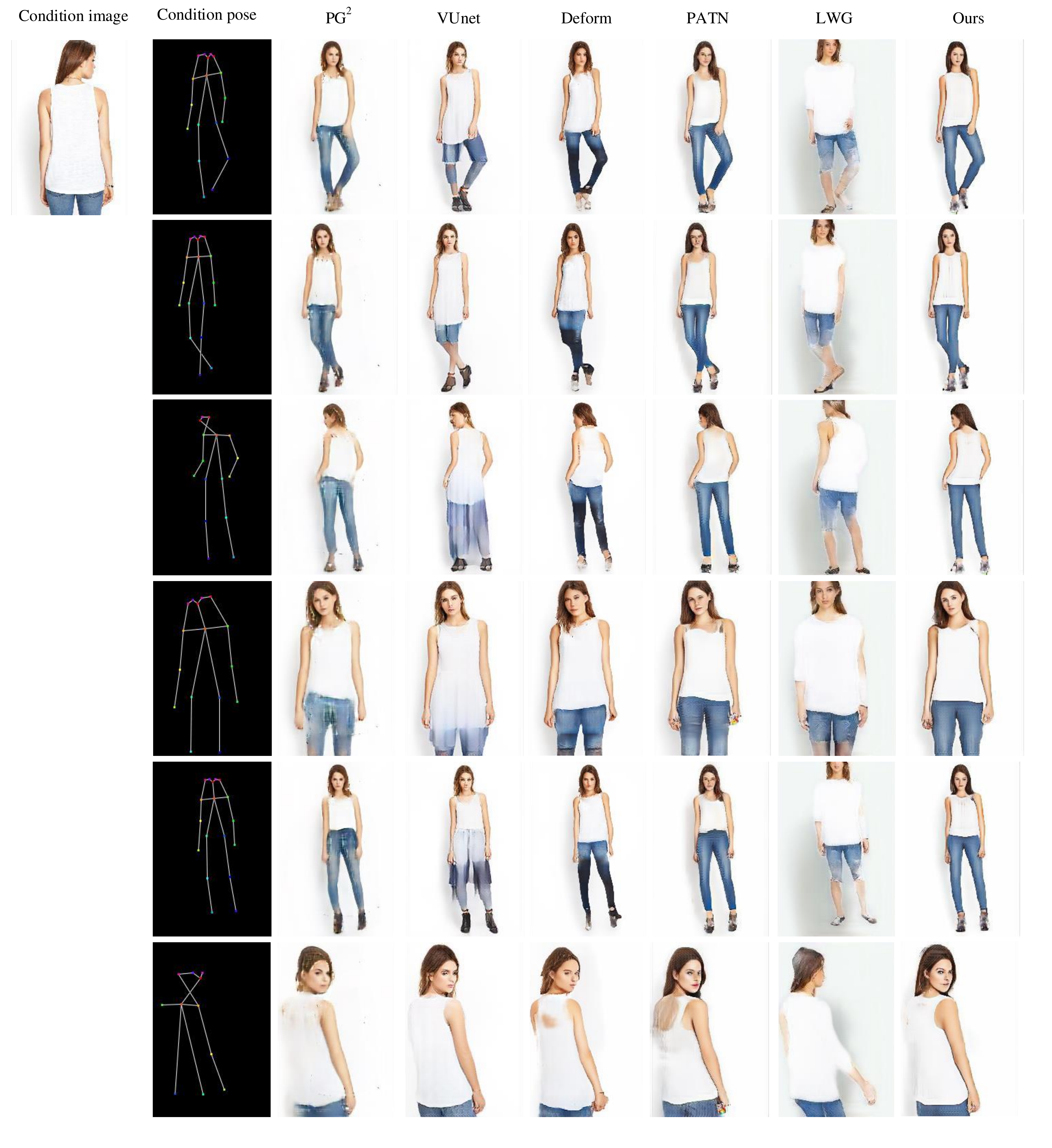}
		\caption{Image generation results conditioned by different poses on the DeepFashion dataset. From left to right are the results of PG$^2$\cite{NIPS2017_6644}, VUnet\cite{Esser_2018_CVPR}, Deform\cite{Siarohin_2018_CVPR}, PATN\cite{Zhu_2019_CVPR}, LWG\cite{lwb2019} and ours, respectively.}
		\label{fig:f_arb1}
	\end{figure*}
	
	To further validate the performance of our model, we try to generate person images based on the same condition image and several different poses from the testing sets of Market-1501 and DeepFashion, respectively. Some examples are shown in Figure \ref{fig:m_arb}-\ref{fig:vslwg1}. We only compare with PATN \cite{Zhu_2019_CVPR} and LWG \cite{lwb2019} since they are the latest state-of-the-art methods and the space is limited. LWG \cite{lwb2019} cannot generate good results on the Market-1501 dataset, and hence we do not compare this method on that dataset. Because LWG \cite{lwb2019} fails to generate sharp image when the condition image and reference image have different camera coordinates, we select some full-body images that have similar camera coordinates as condition images and reference images from DeepFashion dataset. As shown in Figure \ref{fig:m_arb} and Figure \ref{fig:vspatn}, for images in Market-1501 dataset, our model can generate more realistic and plausible images even for poses with large variation. Moreover, our generated images keep shape consistency while the results of PATN \cite{Zhu_2019_CVPR} lose some details and are blurry. Compared with PATN \cite{Zhu_2019_CVPR} on DeepFashion dataset, our model generates sharper images with better facial details and less noise.
As shown in Figure \ref{fig:vslwg1}, the results of LWG are blurry and sensitive to different camera coordinates, although LWG can preserve some details. Besides, due to its dependency on 3D mesh recovery module to get pose and shape information, the persons in the  generated images from different condition images have different shapes and inaccurate pose when the results of 3D mesh recovery module are not precise.
As shown in the third and last columns of the second example, LWG cannot cope well with large differences of camera coordinates between the condition image and the estimated image, which explains why the quantitative results of LWG are not good (see Table \ref{tab:tab_com}).
Besides, we compare with five state-of-the-art methods on DeepFashion dataset in Figure \ref{fig:f_arb1} to show the performance conditioned by different poses for the same person. Our model generates more realistic images and preserves the pattern of clothes. Moreover, the unseen regions synthesized by our model are more reasonable with less artifacts.
		
	\subsubsection{Quantitative Evaluation}
	Table \ref{tab:tab_com} gives quantitative results compared with five state-of-the-art methods: PG$^2$\cite{NIPS2017_6644}, VUnet \cite{Esser_2018_CVPR}, Deform\cite{Siarohin_2018_CVPR}, PATN\cite{Zhu_2019_CVPR} and LWG \cite{lwb2019}.
We use the same training set and testing set used in PATN. Because other methods do not give their data split scheme, we run their well-trained models on our testing set. It is inevitable to have some test images in their training sets. Even in this case, our method still outperforms them on most metrics.
For Market-1501, although our IS metric is slightly lower than Ma \emph{et al.} \cite{NIPS2017_6644}, our mask-IS metric, which is more convincing for Market-1501, is the highest score. The quantitative results on Market-1501 demonstrate that the images generated by our method maintain structure similarity and shape consistency even if the condition images are low resolution and vary significantly in the pose and background. For DeepFashion, our method has the best results in terms of SSIM, which means that our generated images keep structure similarity to the ground truth. Our PSNR value is the highest on both Market-1501 and DeepFashion datasets, which demonstrates that our generated images have the minimum pixel-level errors.
	
	\begin{table}[htbp]
		\renewcommand{\arraystretch}{1.3}
		\small
		\setlength{\tabcolsep}{3mm}
		\begin{center}
			\caption{Comparison of model size and speed on DeepFashion dataset.}\label{tab:tab_eff}
			\begin{tabular}{|c|c|c|}
				\hline
				Method&Parameters & Speed \\
				\hline
				PG$^2$ \cite{NIPS2017_6644} & 437.09 M & 16.36 fps\\
				\hline
				VUnet \cite{Siarohin_2018_CVPR}&  139.36 M & 44.27 fps \\
				\hline
				Deform \cite{Esser_2018_CVPR}& 82.08 M & 25.49 fps\\
				\hline
				PATN  \cite{Zhu_2019_CVPR} & 41.36 M & 97.37 fps \\
				\hline
				LWG \cite{lwb2019} & 97.45 M  & 7.31 fps \\
				\hline
				Ours with 3 PoNA blocks & 34.33 M & 105.18 fps  \\
				\hline
				Ours with 2 PoNA blocks & \textbf{23.31 M} & \textbf{179.32 fps}  \\
				\hline
			\end{tabular}
		\end{center}
	\end{table}

	\subsubsection{Efficiency Evaluation} Table \ref{tab:tab_eff} shows our computation complexity and efficiency compared with five state-of-the-art methods: PG$^2$ \cite{NIPS2017_6644}, VUnet \cite{Esser_2018_CVPR}, Deform \cite{Siarohin_2018_CVPR}, PATN \cite{Zhu_2019_CVPR}, and LWG \cite{lwb2019}. We test all the methods on the same desktop with one NVIDIA Titan Xp graphics card. We discard the time of data preparation, and compute the testing time on DeepFashion dataset. Through the comparison of model size and speed, our model significantly outperforms the other five state-of-the-art methods. Compared with PATN\cite{Zhu_2019_CVPR}, our method has fewer parameters and faster running speed, but get better results, especially for our model with 2 PoNA blocks. Note that the results of our model with 2 PoNA blocks are slightly lower than PATN (with 9 blocks) in SSIM and mask-IS metrics, but higher than their model in IS and mask-SSIM metrics, as shown in Table \ref{tab:tab_com} and Table \ref{tab:num}. In a word, our model has the fewest parameters and the fastest speed, even when using 3 PoNA blocks.

	\subsection{Ablation Study}
	\label{sec:ablation}
	\subsubsection{Effectiveness of Cascaded PoNA Blocks}
\begin{figure*}[!t]
		\begin{center}
			\begin{tabular}{c} 
				\includegraphics[width=0.7\linewidth]{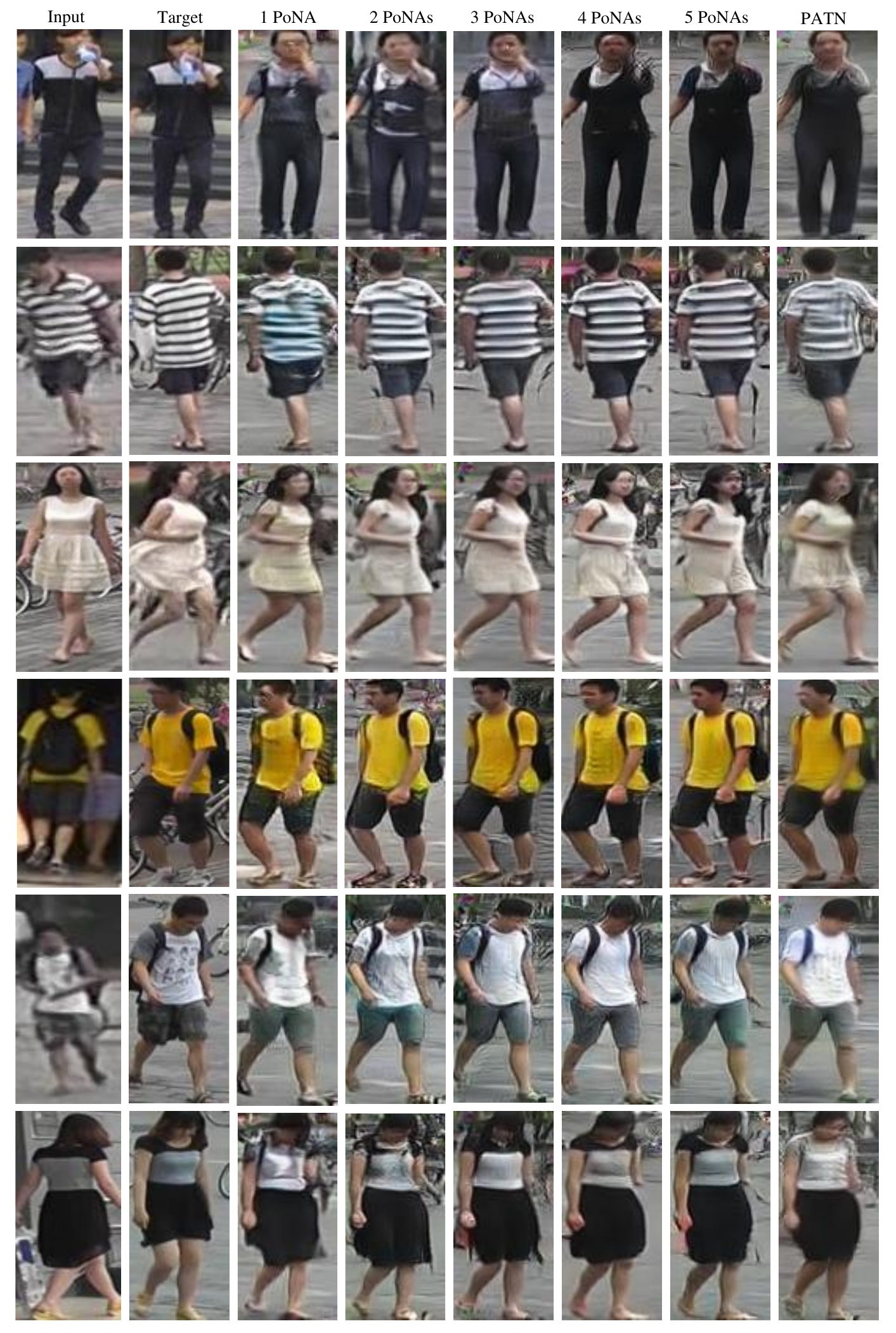}
			\end{tabular}
		\end{center}
		\caption{Results on Market-1501 dataset with different numbers of PoNA blocks.}
		\label{fig:num}
	\end{figure*}

	The generator we proposed has several cascaded PoNA blocks with improved non-local attention mechanism, to deal with the challenging cases of human pose transfer. In order to demonstrate the effectiveness of cascaded PoNA blocks, we test different numbers of PoNA blocks on Market-1501 dataset. Table \ref{tab:tab_eff} and Figure \ref{fig:num} show the quantitative and qualitative results using different numbers of PoNA blocks, respectively. When we only use 1 PoNA block in our model, the quantitative results decrease in all metrics except mask-IS, and slightly lower than the other state-of-the-art methods (please refer to Table \ref{tab:tab_com}). This verifies the effectiveness of our pose-guided non-local attention mechanism. As shown in Figure \ref{fig:num}, even using 1 PoNA block, our model can still obtain reasonable results. For simple clothing patterns, \emph{e.g.}, uniform color skirt (in the third row) and knapsack (in the fourth row), 1 PoNA block is enough. Using 2 or more PoNA blocks, our model can capture fine details of complex appearance and generate plausible images, especially for 3 PoNA blocks. With the increase of PoNA blocks, some artifacts may appear (in the second and third rows). Therefore, in our experiments, we use 3 PoNA blocks to generate plausible and robust results. The images in the last column of Figure \ref{fig:num} are generated by PATN\cite{Zhu_2019_CVPR} with 9 blocks, which has the best performance in their paper. Compared with PATN\cite{Zhu_2019_CVPR}, our PoNA blocks can capture more details and generate more realistic images. With non-local attention mechanism, our network can transfer the details in the condition image, even for the images with complex clothing patterns, and hence our network generates sharper and more realistic images.
	
	\begin{table}[!htbp]
		\renewcommand{\arraystretch}{1.3}
		\small
		\begin{center}
			\caption{Resluts of different numbers of PoNA blocks.}\label{tab:num}
			\begin{tabular}{|c|c|c|c|c|c|}
				\hline
				Number & SSIM & IS &mask-SSIM & mask-IS & PCKh \\
				\hline
				1 & 0.298 & 3.367 & 0.806 & 3.806 & 0.91 \\
				\hline
				2 & 0.309 & 3.419 & 0.812 & 3.745 & 0.94  \\
				\hline
				3 & \textbf{0.315} & \textbf{3.487} & \textbf{0.814} & 3.867 & \textbf{0.94}   \\
				\hline
				4 & 0.309 & 3.481 & 0.809 & 3.783  & 0.92 \\
				\hline
				5 & 0.306 & 3.449 & 0.808 & \textbf{3.872}  & 0.92 \\
				\hline
			\end{tabular}
		\end{center}
	\end{table}

	\subsubsection{Effectiveness of different components of PoNA Block}
\begin{figure}[!ht]
    	\begin{center}
    		\includegraphics[width=0.9\linewidth]{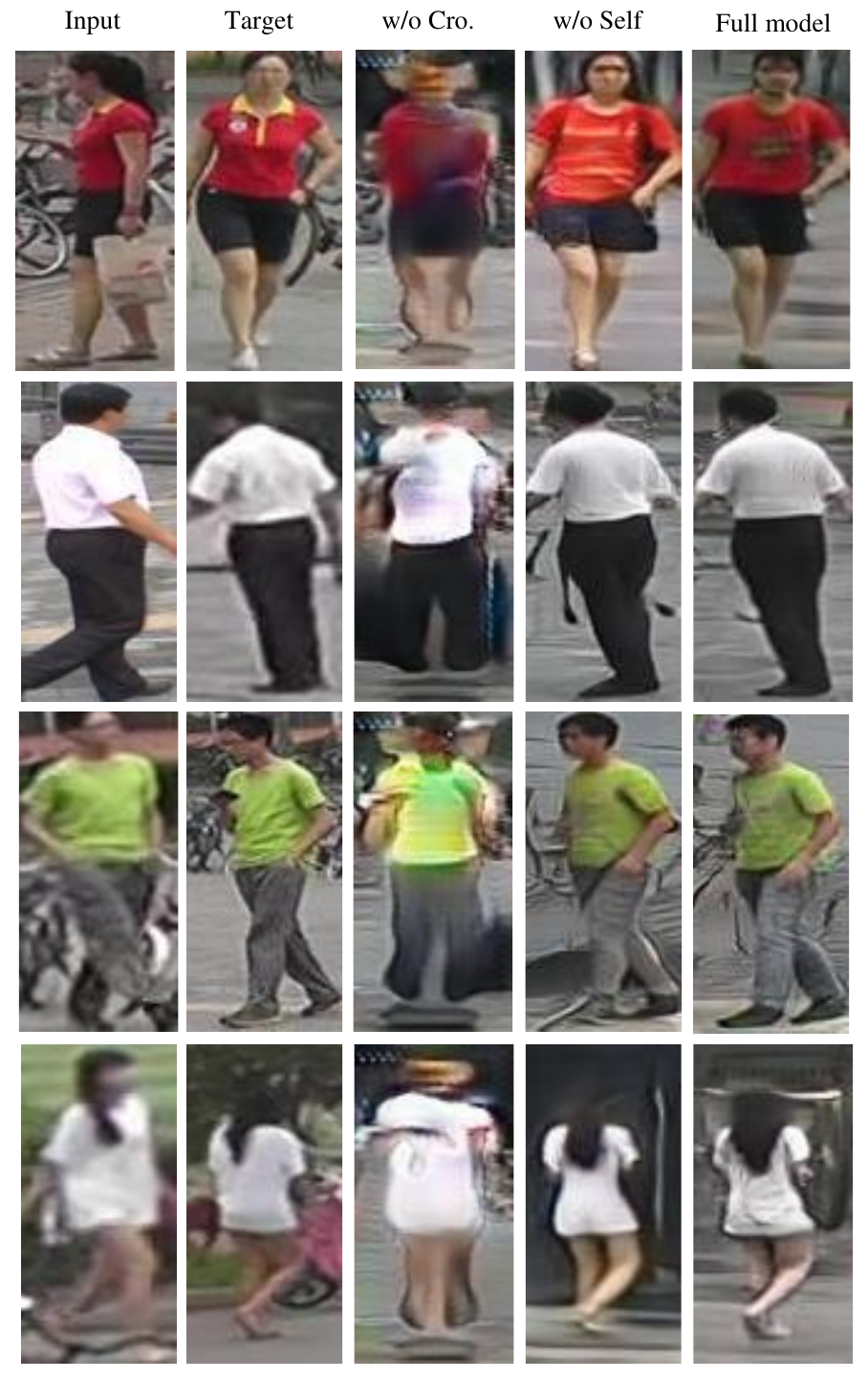}
    	\end{center}
    	\caption{Generated images without different components of the PoNA block.}
    	\label{fig:sc}
    \end{figure}

	The PoNA block is composed of self-attention module and cross-modal attention module. To validate the effectiveness of these two components, we conduct experiments on Market-1501 dataset by training the model without cross-modal module and the model without self-attention module. Table \ref{tab:tab_sc} and Figure \ref{fig:sc} show the quantitative and qualitative results. It can be seen that the cross-modal attention module plays a key role in improving the performance. The cross-modal attention module is designed for post-posed pose-guided image feature update, and it is the only way that pose feature plays a role in guiding the transformation of image feature. Without the cross-modal attention module, our model cannot obtain pose information, which causes that the generated person does not have the similar pose to the ground truth. Besides, there are some different target poses and target images for the same source image as paired training data, which means that the same source image has different target images. Therefore, without pose information, the model may learn to synthesize images with mean shape and mean pose to minimize the loss. In our model, the self-attention module is used to merge pose feature and image feature to select more important features. Without self-attention module, the pre-posed image-guided pose feature update cannot select important features to embed key and value to obtain the attention map, and hence generates some wrong estimates in the final image. Our full model achieves the best performance.

	\begin{table}[!htbp]
		\renewcommand{\arraystretch}{1.3}
		\small
	\setlength{\tabcolsep}{2mm}
		\begin{center}
			\caption{Quantitative results without different components of the PoNA block.}
			\label{tab:tab_sc}
			\begin{tabular}{|c|c|c|c|c|c|}
				\hline
				& SSIM & IS &mask-SSIM & mask-IS & PCKh \\
				\hline
				w/o Cro. & 0.155 & 3.034 & 0.729 & 3.628 & 0.15 \\
				\hline
				w/o Self. & 0.307 & 3.331 & 0.809 & 3.755 & 0.93  \\
				\hline
				Full model & \textbf{0.315} & \textbf{3.487} & \textbf{0.814} & \textbf{3.867} & \textbf{0.94}    \\
				\hline
			\end{tabular}
		\end{center}
	\end{table}

	\subsubsection{Place and Importance of Fusion Code}
\begin{figure}[!t]
		\begin{center}
			\includegraphics[width=1.0\linewidth]{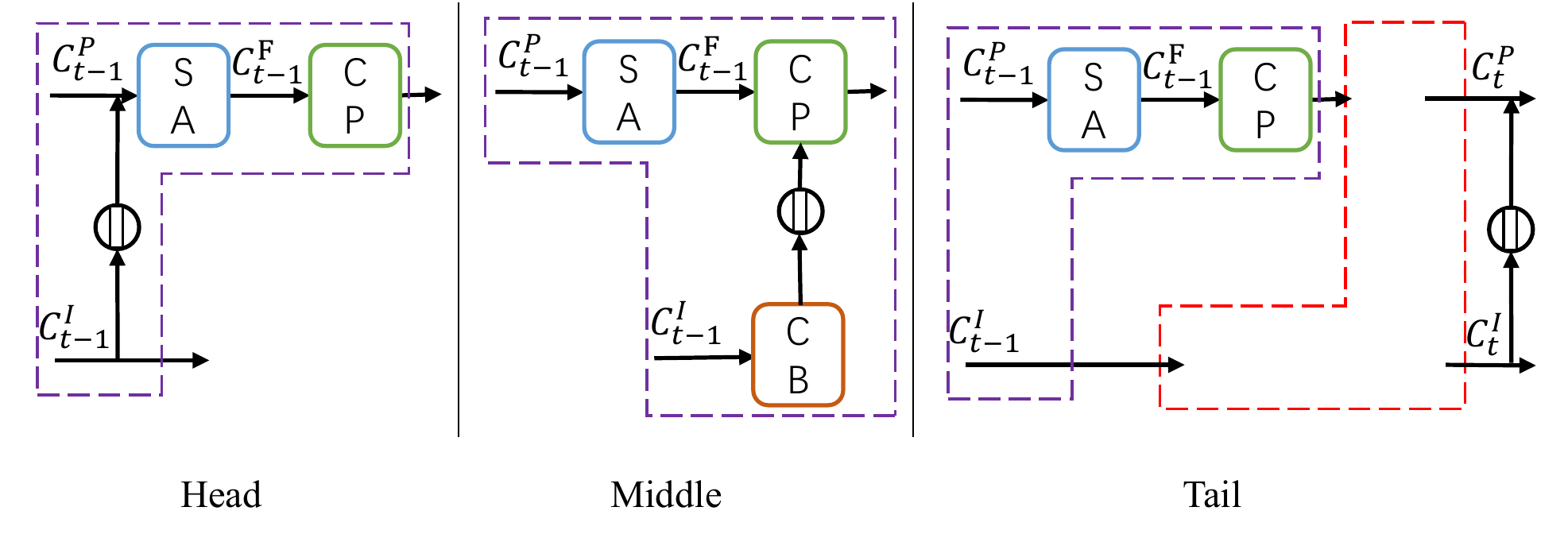}
		\end{center}
		\caption{Structures of different places of fusion code. Note that the legend is as the same as Figure \ref{fig2}. }
		\label{fig:aba_place}
	\end{figure}

	This section will demonstrate our design of the cross-modal block with pre-posed image-guided pose feature update and post-posed pose-guided image feature update.
	The improved attention mechanism can make pose features guide the transfer of image features after getting fusion code and updating pose code, as illustrated in Section \ref{sec:method}. However, because concatenating pose code and image code is the only way to let pose features know the process of image feature transfer, when and where to concatenate image code and pose code to update fusion code is important for the follow-up operations. Images in Market-1501 dataset are more challenging for human pose transfer and suitable for validating the place and importance of fusion code. We call the fusion code before and after pose code updating as the head and tail fusion, and deploy fusion code in the middle of four convolution layers in pose code updating as middle fusion. To validate the importance of fusion code, we also remove it from PoNA block as none fusion. We test our model with 3 PoNA blocks.
	Figure \ref{fig:aba_place} shows the structures of head, middle and tail fusions.

	\begin{table}[!htbp]
		\renewcommand{\arraystretch}{1.3}
		\small
		\begin{center}
			\caption{Results of different places of fusion code.}\label{tab:tab_pla}
			\begin{tabular}{|c|c|c|c|c|c|}
				\hline
				Fusion place & SSIM & IS &mask-SSIM & mask-IS & PCKh \\
				
				\hline
				Head & \textbf{0.315} & 3.487 & \textbf{0.814} & 3.867 & \textbf{0.94} \\
				\hline
				Middle & 0.310 & 3.501 & 0.811 & \textbf{3.897} & 0.93  \\
				\hline
				Tail & 0.307 & \textbf{3.680} & 0.807 & 3.884 & 0.93   \\
				
				\hline
				None & 0.297 & 3.427 & 0.804 & 3.705  & 0.91                      \\
				\hline
				
			\end{tabular}
		\end{center}
	\end{table}

	Table \ref{tab:tab_pla} shows the quantitative results of different places of fusion code. It can be seen that the head fusion, \emph{i.e.}, \emph{pre-posed} image-guided pose feature update, has the best performance, because the head fusion can fuse pose features and image features before pose code updating and help to understand the structured information of image features. The tail fusion cannot maintain the structured information and understand the image features compared with head fusion and middle fusion. Even though we remove the fusion code, owing to our improved attention mechanism, our model can obtain promising results, which are higher or slightly lower than the other state-of-the-art methods illustrated in Table \ref{tab:tab_com} in all metrics. This verifies the effectiveness of our \emph{post-posed} pose-guided image feature update.

\begin{figure*}[!t]
		\begin{center}
			\includegraphics[width=0.7\linewidth]{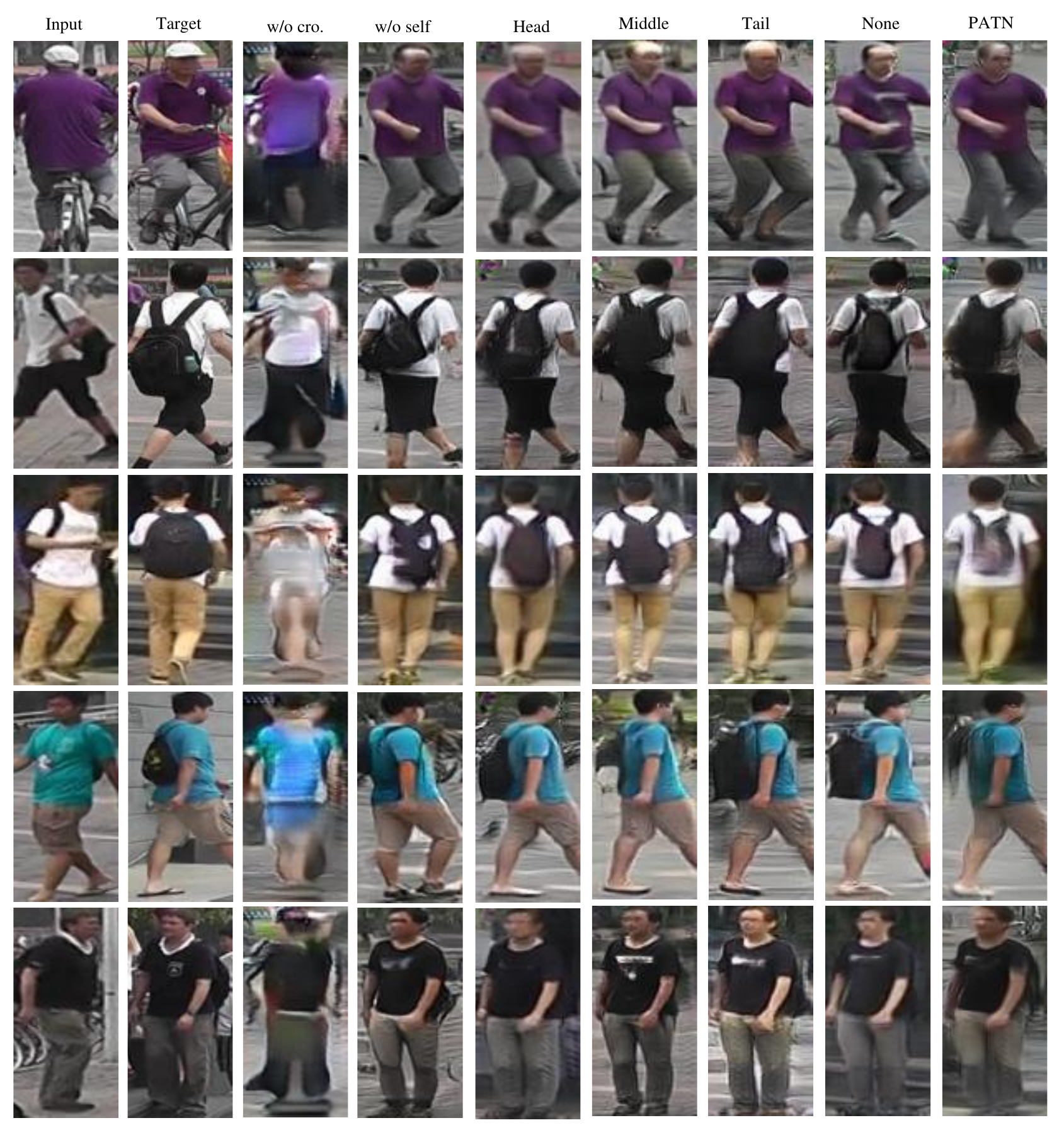}
		\end{center}
		\caption{Generated images using different places of fusion code.}
		\label{fig:pla}
	\end{figure*}

	
	The qualitative results are shown in Figure \ref{fig:pla}. The model with head fusion generates more plausible results based on condition images (in the fourth and fifth rows), especially when the legs in the target image are crossed (in the second row). PATN \cite{Zhu_2019_CVPR} with 9 blocks (in the last column), with local attention mechanism, is not able to capture the details in the condition image, \emph{e.g.}, the white hat (in the first row), the bag (in the third and fourth rows) and the white collar (in the fifth row). Note that the images generated by our model are sharper and have rich details, even if we remove the fusion code updating. This further proves the effectiveness of our improved attention mechanism.

	\subsubsection{Visualization of Features in Our Model}
\begin{figure}[!t]
		\begin{center}
			\includegraphics[width=0.9\linewidth]{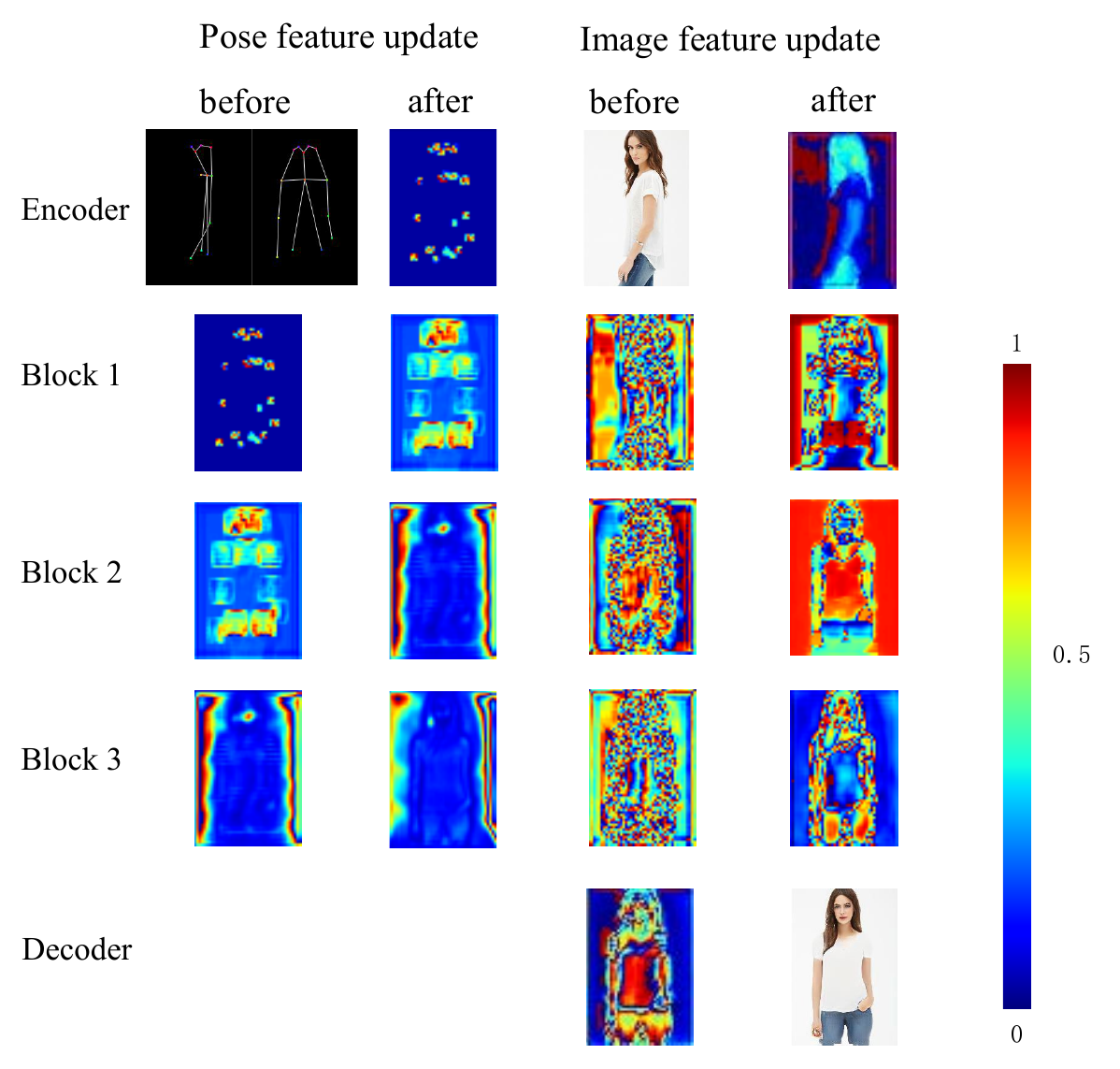}
		\end{center}
		\caption{Visualization of features in our model.}
		\label{fig:att}
	\end{figure}

	In this section, we visualize all the core features of our model to get an intuitive understanding on the transformation of pose feature and image feature. Figure \ref{fig:att} shows the visualization of features in our model. In pre-posed image-guided pose feature update, the images represent the features before and after the encoder and the features before and after the pose code update in each PoNA block from top to down. It can be seen that the pose feature becomes denser, which is beneficial to obtain a reliable attention map to deform the image feature. In post-posed pose-guided image feature update, the first row gives the input image and the output feature of image encoder, and the middle three rows show the image features before and after deforming by cross-modal attention module. As shown in Figure \ref{fig:att}, the cross-modal attention module deforms the image feature from condition pose to target pose gradually and focuses on different regions in each block.

	\subsection{Data Augmentation for Person Re-identification}
	\label{sec:aug}
	Human pose transfer is able to generate images of the same person in different poses, which is useful to augment person re-identification (re-ID) \cite{zheng2016person} dataset to solve the problem of lacking training data and improve the performance of person re-ID. To some degree, the performance of augmenting dataset depends on the performance of human pose transfer model. In order to illustrate the performance of our model, we test on Market-1501 dataset \cite{zheng2015scalable}, which is a main person re-ID dataset. We exploit our generated images to replace the images in Market-1501. Specifically, we first obtain a reduced training set by selecting a portion \emph{p} to randomly preserve the images in the whole training set. Then, we use our human pose transfer method to generate missing images conditioned on preserved images and the pose of missing images. Finally, we combine the reduced training set and the generated training set to obtain the new training set. Note that the images in the new training set has the same identities and each identity has the same number of images with the same pose as original images. We select the portion from $20\%$ to $80\%$ at intervals of $20\%$, and get four reduced training sets and four new augmented training sets. We use the Person re-ID baseline Framework \cite{zheng2019joint} based on ResNet-50 \cite{resnet} as our training and testing protocols.

	The training sets we created are suitable for showing the performance of data augmentation using our human pose transfer method. First, the reduced training set provides an environment with insufficient data and helps us know the lower bound of performance. Second, the original training set with realistic images gives the upper bound of performance. With lover bound and upper bound, we can measure the performance of data augmentation using the human pose transfer method.
Table \ref{tab:reid} shows the re-ID results using the reduced training sets (referred to as None) and the new training sets generated by VUnet\cite{Esser_2018_CVPR}, Deform\cite{Siarohin_2018_CVPR}, PATN\cite{Zhu_2019_CVPR}, and our method. We did not compare PG$^2$\cite{NIPS2017_6644} and LWG\cite{lwb2019} since they cannot generate good results on this Market-1501 dataset. For fair comparison, we use the same settings (\emph{e.g.}, condition images and target poses) to generate the images for all the methods.
We use mean Average Precision (mAP) as the metric to measure the re-ID performance.
With the same model and the same parameters, the re-ID performance relies on the photorealism of generated images and the texture consistency of the same identity. As shown in the table, the model augmented by our method achieves the most accurate re-ID estimation, which indicates that our method generates more realistic images with consistent textures.

\begin{table}[!h]
	\renewcommand{\arraystretch}{1.3}
	\small
	\setlength{\tabcolsep}{3.5mm}
	\begin{center}
		\caption{Results of re-ID.}
		\label{tab:reid}
		\begin{tabular}{|c|c|c|c|c|c|}
			\hline
			\multirow{2}*{Aug. Model} &\multicolumn{5}{|c|}{Portion \emph{p} of the original dataset}   \\ \cline{2-6}
			{}& 0.2 & 0.4  & 0.6  & 0.8 &1.0   \\
			\hline
			None&42.3&58.9&66.7&69.5&71.8   \\
			VUnet\cite{Esser_2018_CVPR} & 54.1 & 59.6 & 66.2 & 68.6 & 71.8  \\
			Deform\cite{Siarohin_2018_CVPR} & 55.8 & 60.7 & 67.4 & 69.4 & 71.8    \\
			PATN\cite{Zhu_2019_CVPR}&57.2 & 61.4& 67.9 & 69.7 & 71.8     \\
			\hline
			Ours& \textbf{58.8} & \textbf{63.8} & \textbf{68.7} & \textbf{70.4} & \textbf{71.8}   \\
			\hline
			
		\end{tabular}
	\end{center}
\end{table}

	
	\subsection{Limitations}
\label{sec:limitation}
\begin{figure}[!ht]
	\centering
	\includegraphics[width=1.0\linewidth]{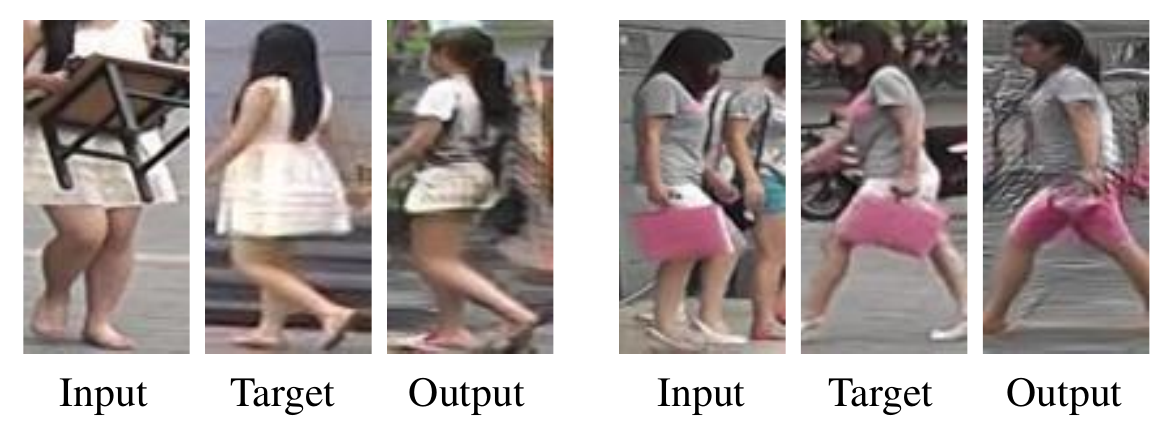}
	\caption{Examples of failure cases using our method.}
	\label{fig:fail}
\end{figure}

	Our model can deform the image feature from the condition pose to the target pose by pose-guided attention mechanism, which also alleviates the negative effects of occlusion. Although our model generates impressive results, the quality of generated images can be further improved, especially for occlusion cases. Figure \ref{fig:fail} shows some failure cases using our method.
Our model treats large areas of occlusion as part of the texture, resulting in blurry areas and incorrect textures (\emph{e.g.}, dress and shorts). Pose-guided attention mechanism, as a non-local attention mechanism, can select and deform important regions, but cannot cope well with the invisible areas in the condition image by the weighted sum. In future work, we will try to use human parsing map to extract semantic information to deal with occlusion and add local attention mechanism to enhance the quality of texture.

	\section{Conclusion}
	\label{sec:conclude}
	In this paper, we propose a pose-guided non-local attention (PoNA) block with an improved attention mechanism to deal with the challenging human pose transfer. With the improved attention mechanism, each block selects precise regions of image features to transfer based on pose features. The generator of our network is composed of several PoNA blocks and transfers image features progressively. Compared with previous work, our network generates more realistic and sharper images with rich details, and get the highest scores in the reasonable mask-SSIM and mask-IS metrics. At the same time, our network has fewer parameters and faster speed. Moreover, the proposed network can generate training images for person re-identification to alleviate the data insufficiency. Our improved attention mechanism with pre-posed and post-posed fusion is suitable for other conditioned generation tasks. In the future, we will deploy the improved attention mechanism to other conditioned generation tasks, such as facial animation.


	
	
	\bibliographystyle{IEEEtran}
	\bibliography{ref}

	%
	
	
	
	%

	
	

\end{document}